# A Comprehensive Survey of Grammar Error Correction

Yu Wang*, Yuelin Wang*, Jie Liu[i], and Zhuo Liu

*Abstract*—**Grammar error correction (GEC) is an important application aspect of natural language processing techniques. The past decade has witnessed significant progress achieved in GEC for the sake of increasing popularity of machine learning and deep learning, especially in late 2010s when near human-level GEC systems are available. However, there is no prior work focusing on the whole recapitulation of the progress. We present the first survey in GEC for a comprehensive retrospect of the literature in this area. We first give the introduction of five public datasets, data annotation schema, two important shared tasks and four standard evaluation metrics. More importantly, we discuss four kinds of basic approaches, including statistical machine translation based approach, neural machine translation based approach, classification based approach and language model based approach, six commonly applied performance boosting techniques for GEC systems and two data augmentation methods. Since GEC is typically viewed as a sister task of machine translation, many GEC systems are based on neural machine translation (NMT) approaches, where the neural sequence-to-sequence model is applied. Similarly, some performance boosting techniques are adapted from machine translation and are successfully combined with GEC systems for enhancement on the final performance. Furthermore, we conduct an analysis in level of basic approaches, performance boosting techniques and integrated GEC systems based on their experiment results respectively for more clear patterns and conclusions. Finally, we discuss five prospective directions for future GEC researches.**

**Be sure that you adhere to these limits; otherwise, you will need to edit your abstract accordingly. The abstract must be written as one paragraph, and should not contain displayed mathematical equations or tabular material. The abstract should include three or four different keywords or phrases, as this will help readers to find it. It is important to avoid over-repetition of such phrases as this can result in a page being rejected by search engines. Ensure that your abstract reads well and is grammatically correct.**

## I. INTRODUCTION

ENGLISH boasts the biggest number of speakers around the world. For most of English speakers, English is not their natural language, thus they are under insufficient language proficiency level and are more inclined to make grammar errors. Their expressions could be corrupted by the noise in the form of interference from their first-language background and therefore contains error patterns dissimilar to that in essays of native speakers. Building a system that automatically corrects grammar errors for English learners becomes increasingly necessary, which could be applied in many scenarios, such as when people are writing essays, papers, statements, news and emails. In this situation, researches about developing grammar error correction systems have received more and more attention and much progress has been achieved.

Grammar error correction (GEC) aims for automatically correcting various types of errors in the given text. Errors that violate rules of English and expectation usage of English native speakers are in morphological, lexical, syntactic and semantic forms are all treated as target to be corrected. Most of GEC systems receive a raw, ungrammatical sentence as input, and return a refined, correct sentence as output. A typical instance is shown in Figure 1.

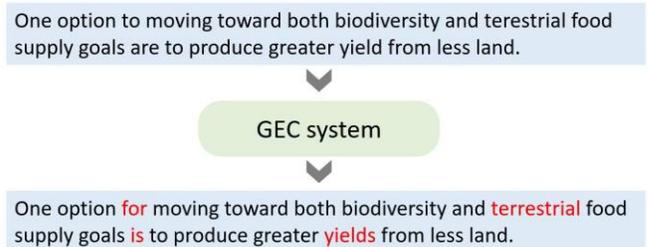

Fig. 1. A typical instance of grammar error correction.

The 2000s is the prior stage for GEC. In this stage, most of GEC systems are based on hand-craft rules incorporating usage of parsers and linguistic

• Y. Wang, Y. Wang, J. Liu and Z. Liu are with College of Artificial Intelligence, Nankai University, Tianjin, China, 300350.
• E-mail: {yuwang17@mail., yuelinwang17@mail., jliu@}nankai.edu.cn, zhuoliu1711475@gmail.com

*: Equal contribution.
i: Corresponding author.

characteristics, for example, the Language Tool [1] and the ESL Assistant [2]. However, the complexity of designing rules and solving conflicts among the rules require a great magnitude of labour. Although some GEC works today still use rules as an additional source of correction, the performance of rule based GEC systems has been superseded by data-driven approaches, which are the focus of our survey.

Remarkable progress has been achieved in GEC by data-driven approaches. In late 2000s, classification based approaches are developed to correct preposition errors and article errors. In this approach, classifiers are trained on large magnitude of native error-free text to predict the correct target word, taking account into the linguistic features given by context. However, GEC systems that are capable of correcting all types of errors are more desirable. As an improvement, statistical machine translation (SMT) based GEC systems have gained significant attention in the early 2010s. In this approach, SMT models are trained on parallel sentence pairs, correcting all types of errors by "translating" the ungrammatical sentences into refined output. More recently, with the increasing popularity of deep learning, neural machine translation (NMT) based GEC systems applying neural seq2seq models become dominant and achieve the state-of-the-art performance. Since MT based translation approaches require corpus containing large parallel sentence pairs, language model based GEC approaches have been researched as an alternative that does not rely on supervised training. We give each of the data-driven approaches in Section 3 in detail.

Beyond the four basic approaches, numerous techniques have also attracted significant attention in order to facilitate the GEC system to achieve better overall performance, especially in SMT and NMT based GEC systems. Since GEC is an area that stresses the appropriate application of basic model, the techniques are important to adapt the existing models to GEC. These techniques have been explored and developed incrementally to provide assistance and could be combined to further improve the error correction ability of GEC systems. In our survey, we describe the researches about GEC techniques and their broad application in existing works.

Data augmentation methods are also of great essence to development of GEC. The lack of large amounts of public training sentence pairs refrains the development of more powerful MT based GEC systems. This problem could be partly ameliorated by the proposal of various data augmentation methods, which generate artificial parallel data for training GEC models. Some inject noise into error-free texts for corruption, while others apply back translation on error-free texts to translate them into ungrammatical counterparts. Both methods are commonly applied in today's GEC systems.

Apart from the aforementioned components of GEC systems, we also cover other aspects of the GEC task. We summarize the statistics and properties of several public GEC datasets, briefly discuss the researches about data annotation schema, and introduce two GEC shared tasks, which are critical to the development of GEC. Besides, standard evaluation provides a platform where multiple GEC systems could be compared and analysed quantitatively. The evaluation metrics, including both reference-based and reference-less, are explained and compared.

GEC has always been a challenging task in NLP research community. First, due to the unrestricted mutability of language, it is hard to design a model that is capable of correcting all possible errors made by non-native learners, especially when error patterns in new text are not observed in training data. Second, unlike machine translation where annotated training resources are abundant, a large amount of annotated ungrammatical texts and their corrected counterparts are not available, adding difficulties to training MT based GEC models. Although data augmentation methods are proposed to alleviate the problem, however, if the artificially generated data cannot precisely capture the error distribution in real erroneous data, the final performance of GEC systems will be impaired.

We provide a comprehensive literature retrospective on the research of GEC. The overall categorization of the research is visualized in Figure 2.

Our survey makes the following contributions.

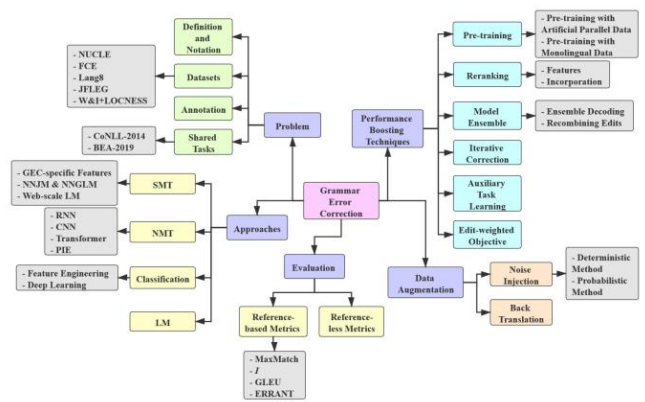

Fig. 2. Categorization of research on GEC.

· We present the first comprehensive survey in GEC. To the best of our knowledge, no prior work focuses on the overall survey in all aspects of datasets, approaches, performance boosting techniques, data augmentation methods and evaluation of the great magnitude of researches in GEC, especially the explorations in recent years that yield significant progress.

· We explicitly separate the elements belonging to the approaches, performance boosting techniques and data augmentation and put them together for a more structured description of GEC works. Due to the nature of GEC, this is more beneficial for following works based on the incorporation and application of disparate approaches, techniques and data augmentation methods.

TABLE 1
Statistics and properties of public GEC datasets.

| Corpus | Component | # Sents | # Tokens | # Chars per sent | Sents Changed | # Ref | Error Type | Error Type | Proficiency | Topic | L1 |
|--------|-----------|---------|----------|------------------|---------------|-------|------------|------------|-------------|-------|-----|
| NUCLE | - | 57k | 1.16M | 115 | 38% | 2 | minimal | Labeled | Simplex | Simplex | Simplex |
| FCE | Train | 28k | 455k | 74 | 62% | 1 | minimal | Labeled | Simplex | Diverse | Diverse |
| | Dev | 2.1k | 35k | | | | | | | | |
| | Test | 2.7k | 42k | | | | | | | | |
| Lang-8 | - | 1.04 M | 11.86 M | 56 | 42% | 1-8 | fluency | None | Diverse | Diverse | Diverse |
| JFLEG | Dev | 754 | 14k | 94 | 86 | 4 | fluency | None | Diverse | Diverse | Diverse |
| | Test | 747 | 13k | | | | | | | | |
| W&I | Train | 34.3k | 628.7k | 60 | 67% | 1 | - | Labeled | Diverse | Diverse | Diverse |
| | Dev | 3.4k | 63.9k | 94 | 69% | 1 | | | | | |
| | Test | 3.5k | 62.5k | - | - | 5 | | | | | |
| LOCNESS | Dev | 1k | 23.1k | 123 | 52% | 1 | - | Labeled | Diverse | Diverse | Simplex |
| | Test | 1k | 23.1k | - | - | 5 | | | | | |

- We make a recapitulation of current progress and present an analysis based on empirical results in aspects of the approaches, performance boosting techniques and integrated GEC systems for a more clear pattern of existing literatures in GEC.
- We propose five prospective directions for GEC in aspects of adapting GEC to native language of English learners, low-resource scenario GEC, combination of multiple GEC systems, datasets and better evaluation based on the existing works and progress.

The rest of this survey is organized as follows. In Section 2, we introduce the public GEC corpora and the annotation schema. Section 3 summarizes the commonly adopted data-driven GEC approaches. We classify the numerous techniques in Section 4 and collect data augmentation methods in Section 5. The standard evaluation metrics and discussion of GEC systems are in Section 6. In Section 7, we propose the prospective directions and conclude the survey in Section 8.

## II. PROBLEM

In this section, we introduce the fundamental concepts of GEC and their notations in the survey, the public datasets, the annotation schema of data and the shared tasks in GEC.

### 2.1 Notation and Definition

Since GEC is an area of application of disparate theories and models, we focus only on the universal concepts that are shared among the most GEC systems.

Given a source sentence $x = \{x_1, x_2, ..., x_{T_x}\}$ which contains grammar errors, a GEC system learns to map x to its corresponding target sentence $y = \{y_1, y_2, ..., y_{T_y}\}$ which is error-free. The output of the GEC system is called hypothesis $\hat{y} = \{\hat{y}_1, \hat{y}_2, ..., \hat{y}_{T_{\hat{y}}}\}$.

### 2.2 Datasets

We first introduce several datasets that are most widely used to develop GEC systems based on supervised approach, including NUCLE [3], Lang-8 [4], FCE [5], JFLEG [6], Write&Improve+LOCNESS [7]. These datasets contain parallel sentence pairs that are used to develop MT based systems. Statistics and more properties of these datasets are listed in Table 1. Then, we also describe some commonly used monolingual corpora.

#### 2.2.1 NUCLE

The NUS Corpus of Learner English (NUCLE) is the first GEC dataset that is freely available for research purposes. NUCLE consists of 1414 essays written by Asian undergraduate students at the National University of Singapore. This leads to low diversity of topic, sentence proficiency and L1 (the first language of writer) of the data. Most tokens in NUCLE are grammatically correct, resulting 46,597 annotated errors for 1,220,257 tokens. NUCLE is also annotated with error types.

#### 2.2.2 FCE

The First Certificate in English Corpus (FCE) is a portion of proprietary Cambridge Learner Corpus (CLC) [8], and it is a collection of 1244 scripts written by English language learners in respond to FCE exam questions. FCE contains broader native languages of writers, and more diverse sentence proficiency and topic. Similar to NUCLE, FCE is also annotated with error types. However, FCE is annotated by only 1 annotator, which may lead to fewer usages than NUCLE.

### 2.2.3 Lang-8

The Lang-8 Corpus of Learner English (Lang-8) is a somewhat-clean, English subsection of a social language learning website where essays are posted by language learners and corrected by native speakers. Although Lang-8 contains the maximal amount of original sentences, it is corrected by users with different English proficiency, weakening the quality of the data. Besides, it is not annotated with error types.

### 2.2.4 JFLEG

The JHU FLuency-Extended GUG Corpus (JFLEG) contains 1511 sentences in GUG development and test set. Unlike NUCLE and FCE, annotation of JFLEG involves not only grammar error correction but also sentence-level rewrite to make source sentence sound more fluent. Each sentence is annotated four times on Amazon Mechanical Turk. JFLEG provides new perspective that GEC should make fluency edits to source sentences.

### 2.2.5 W&I+LOCNESS

Write&Improve Corpus (W&I) and LOCNESS Corpus are recently introduced by The BEA-2019 Shared Task on Grammatical Error Correction. W&I consists of 3600 annotated essay submissions from Write&Improve [9], an online web platform that assists non-native English students with their writing. The 3600 submissions are distributed in train set, development set and test set with 3000, 300 and 300 respectively. LOCNESS Corpus is a collection of about 400 essays written by British and American undergraduates, thus it contains only native grammatical errors. Note that LOCNESS contains only development set and test set. All the sentences in W&I and LOCNESS are evenly distributed at each CERF level.

### 2.2.6 Monolingual Corpora

Except for the parallel datasets discussed above, there are also some public monolingual corpora that can be used in training language models, pre-training neural GEC models and generating artificial training data. The commonly used monolingual corpora are listed below.

- **Wikipedia.** Wikipedia is an online encyclopedia based on Wiki technology, written in multiple languages with more than 47 million pages.
- **Simple Wiki.** The Simple English Wikipedia, compared to ordinary English Wikipedia, only uses around 1500 common English words, which makes information much easier to understand both in grammar and structure.
- **Gigaword.** The English Gigaword Corpus is collected by the Linguistic Data Consortium at the University of Pennsylvania, consisting of comprehensive news text data (including text from Xinhua News Agency, New York Times, etc.).
- **One Billion Word Benchmark.** The One Billion Word Benchmark is a corpus with more than one billion words of training data, aiming for measuring research progress. Since the training data is from the web, different techniques can be compared fairly on it.
- **COCA.** The Corpus of Contemporary American English is the largest genre-balanced corpus, containing more than 560 million words, covering different gen-res including spoken, fiction, newspapers, popular magazines and academic journals.
- **Common Crawl.** The Common Crawl corpus [10] is a repository of web crawl data which is open to everyone. It completes crawls monthly since 2011.
- **EVP.** The English Vocabulary Profile is an online extensive research based on CLC, providing information, including example sentences of words and phrases for learners at each CERF level.

## 2.3 Annotation

In this section, we talk about the annotation of GEC data, including researches about annotation schema and inter-annotator agreement. Although annotation requires a great magnitude of labor and time, it is of great importance to the development of GEC. Data-driven approaches rely heavily on annotated data. Most importantly, it enables comparable and quantitative evaluation of different GEC systems when they are evaluated on data with standard annotation. For example, systems in the CoNLL-2014 and the BEA-2019 shared tasks are evaluated on data with official annotation, and they are ranked according to scores using identical evaluation metrics respectively. Besides, it also promotes targeted evaluation and strength of GEC systems, since error types may be annotated in data. We can examine the performances of GEC systems on specific error types to identify their strength and weakness. This is meaningful since a robust specialised system is more desirable than a mediocre general system [11].

The most widely applied annotation is error-coded. Error-coded annotation involves identifying (1) the span of grammatical erroneous context (2) error type and (3) corresponding correction. Many error-coded annotation schemas are applied, but M2 format is most commonly used error-coded annotation format today. M2 format can merge an-notations from multiple annotators, as shown in Figure 3.

```
S Surrounded by such concerns , it is very
likely that we are distracted to worry about
these problems .
A 13 14|||Trans|||and|||REQUIRED|||-NONE-|||0
A 11 12|||Vt|||will be|||REQUIRED|||-NONE-|||1
A 12 12|||Wform|||too|||REQUIRED|||-NONE-|||1
```

Fig. 3. Example M2 annotation format. Line preceded by *S* represents the original sentence, while line preceded by *A* represents an annotation, which consists of start token offset, end token offset, error type, correction and some additional information. Note that the last number of each annotation line denotes annotator's ID.

Error-coded annotation has some disadvantages. First, different corpora classify error types with significant discrepancy. As mentioned above, error types in FCE are distinguished into 80 categories, while in NUCLE are distinguished into only 27 categories. Besides, annotation may vary with different annotators due to their different linguistic background and proficiency, resulting into low inter-annotator agreement (IAA) [12], [13]. Non-error-coded fluent edits which treat correcting sentence as whole-sentence fluency boosting rewriting may be more desirable [14].

There are also many researches on inter-annotator agreement and annotation bias. Annotation bias is a tricky problem in GEC that different annotators show bias towards specific error type and multiple corrections. However, annotation bias can be partly overcome by increasing the number of annotators [15], [16]. It is also demonstrated that by taking increasing number of annotation as golden standard, system performance also improves [17].

## 2.4 Shared Tasks

The shared tasks of GEC have made great contribution to the development of GEC researches. Participating teams are encouraged to submit their error correction systems for higher testing scores, during which many advances, especially in techniques, have been made. While HOO-2011 [18], HOO-2012 [19], CoNLL-2013 [20] shared tasks focus on several specific error types, teams participated in the CoNLL-2014 [21] and BEA-2019 [7] shared tasks are required to correct all types of errors. We briefly discuss CoNLL-2014 and BEA-2019 shared task below. The adopted evaluation metrics in the two tasks are discussed in Section 6.

### 2.4.1 CoNLL-2014

The CoNLL-2014 shared task is the first GEC shared task that aims at correcting all types of errors, which is divided into 28 classifications. The training set is processed NUCLE dataset, while the test set is composed of 50 essays written by non-native English students. The test data is annotated by 2 experts, and 8 more annotations are released afterwards [17]. The official evaluation metric is $M^2F_{0.5}$ score. It is worthy to mention that CoNLL-2014 test set is the most widely used benchmark test set in today's GEC researches.

### 2.4.2 BEA-2019

The BEA-2019 shared task reevaluates the more developed and various GEC systems in a unified condition 5 years after the CoNLL-2014. One contribution of this task is the introduction of a new parallel GEC dataset W&I+LOCNESS. ERRANT is used to standardize the edits in several corpora and classify the errors into 25 categories. The test set of W&I+LOCNESS consists of 4477 sentences. System outputs are evaluated in ERRANT $F_{0.5}$ score. There are 3 tracks in the BEA-2019 shared task. In the restricted task, participants can only use NUCLE, FCE, Lang-8 and W&I+LOCNESS as annotated training source, but the monolingual corpus is not restricted. In the unrestricted track, participants can use private datasets or external resources to build their GEC systems. In low-resource track, participants can only use W&I+LOCNESS.

## III. Approaches

In this section, we survey the basic data-driven approaches used to tackle GEC problems, including SMT based approaches, NMT based approaches, classification based approaches and LM based approaches. We divide this section in several branches to clarify the development of various GEC approaches individually, and each branch includes more details about the representative works.

## 3.1 SMT Based Approaches

Before resorting to SMT, grammatical error corrections were mainly achieved by rule based or classification based methods, the limitations of which are obvious. On the one hand, designing rules often requires a large amount of prior linguistic information and expert knowledge. The progress of designing rules is time-consuming and the linguistic resources do not appear to be available all the time, especially for some minority languages that are spoken by only a few people. At the same time, an one-for-all rule is almost non-existent, because the natural language itself is so flexible that makes it difficult for rules to effectively deal with all possible exceptions. On the other hand, the flexibility of natural language also restraints classification based methods to achieve considerable results under a more extensive scene. The preset labels limit the classification method to certain types of errors and cannot be extended to deal with more complicated error types. For GEC, generally speaking, the errors appeared often involve not only the wrong words themselves, but also the context which may contain the surrounding tokens in a sentence or even cross sentence information. Further, the choice of correction is also diverse and flexible, which promotes the development of generative model on GEC. In this section, we present SMT based models and its development on grammatical error correction. Since most current researchers focus on models based on neural machine translation, we do not discuss too many particulars of models and methods in detail here.

### 3.1.1 Statistical Machine Translation

Machine translation aims to find a translation $y$ in target language for a given source sentence x in the source language, which probabilistically finds a sentence y that maximize $p(y|x)$. The distribution $p(y|x)$ is the translation model we want to estimate. Often, the translation model consists of two parts, the translation probability $\Pi_{i=1}^I p(u_i|v_i)$ of the all "units" in a sentence with $U$ units and the reordering probability $\Pi_{i=1}^U d(start_i - end_{i-1} - 1)$, which are used as the basis for translation in different models:

$$\arg\max_y p(y|x) = \arg\max_y \prod_{i=1}^U p(u_i|v_i) d(start_i - b_{end-1} - 1) \quad (1)$$

where $u_i$ and $v_i$ represent the basic "units" in target and source sentence respectively. Similarly, $start_i$ and $end_i$ represent the positions of the first and last token inside the $i$-th unit in the sentence. The contents of "units" depend on the specific model people selected. The word-based translation model uses words as the basic unit, while the phrase-based model uses phrases as the basic unit, and correspondingly, the tree-based method establishes the probability of mapping subtrees between the source language syntax tree and the target language syntax tree. Using Bayes' formula, we can transform the above formula into the noisy channel model:

$$\arg\max_y p(y|x) = \arg\max_y p(x|y)p(y). \quad (2)$$

In this model, it is noted that in addition to an inverted translation model, it also includes a language model $p(y)$ on the target side. By adding a language model, the fluency of the output sentence can be scored and improved. These two models are independent, and training a language model does not require parallel corpora. A monolingual corpus with a large amount of target language can obtain a good language model. The introduction of the noise channel model allows us to first train the language model and the translation model separately and then uses the above formula to combine the two models. This idea is better reflected on the log-linear model, which generalizes it with the following equation:

$$\hat{y} = \arg\max_y p(y|x) = \arg\max_y \exp(\sum_{m=1}^M \lambda_m f_m(x, y)). \quad (3)$$

Compared to the noise channel model, the log-linear model provides a more general framework. This framework contains a variable number of sub-models, namely the feature functions $f_m = (x, y)$, which could be features such as translation model or language model, and a group of tunable weight parameters $\lambda_m$, which are used to adjust the effect of each sub-model in the translation process. Also, the total number of feature functions is represented as $M$. With the log-linear model, prior knowledge and semantic information are possible to be injected by adding feature functions.

SMT based methods treat GEC as the translation process from the "bad English" to "good English". Through training on a large number of parallel corpora, the translation model can collect the corresponding between the grammar error and its correct form, just as in the translation process does in mapping bilingual sentences. Several researches have used statistical machine translation models as the basic framework for GEC, and taken the advantage of the versatility and flexibility provided by log-linear models with considerable result achieved.

### 3.1.2 SMT Based Methods in GEC

The pilot study using SMT for grammatical error correction focused on correcting countability errors of mass noun [22]. A dependency treelet-based SMT model has been employed with artificially constructed parallel corpus as a result of the lack of parallel data required by SMT. By testing at the web-crawled CLEC dataset, their model achieved 61.52% accuracy in correcting mass noun errors and show the promising application of SMT, under the assistance of parallel GEC data, to solve more general errors. Later, the potential competence of SMT based methods were further explored on unrestricted error types [23], [24] through large-scale error corrected data from Lang-8. Also, in the CoNLL-2013 and CoNLL-2014 shared tasks, several SMT based approaches have been employed as significant roles in those top systems [25], [26], [27], [28] with considerable performance achieved. Among those works, task-specific features and web-scale language models were incorporated into standard SMT models [27], [28] and with appropriate tuning approach, their systems ranked first and third respectively on the CoNLL-2014 test set.

Work [29] after those shared tasks also demonstrated the effectiveness of systems utilizing SMT by constructing a hybrid system that combines the outputs of both the SMT components and the classification approach. In this research, four systems (two SMT based systems and two classification based systems respectively) are combined using MEMT [30] and the final system achieves a state-of-the-art result. Another commonly used feature which is based on neural networks was first integrated in SMT based GEC system in more recent studies [31], both a neural network joint model (NNJM) and a neural network global lexical model (NNGLM) are introduced into the SMT based GEC models. The NNJM is further improved using the regularized adaptive training method described in later work [32] on a higher quality training dataset, which has a higher error per-sentence ratio.

TABLE 2
A brief summary of SMT based systems in GEC concerning a range of key properties widely employed in those approaches. The column "Training LM" indicates different monolingual corpora used to train the language models. Also often the target side of parallel corpus is sorted out as monolingual training data for LM, which we omit for simplicity.

| Source | GEC-Specific Features | | Reranking | | Other Attributes | Training LM | Year |
|---|---|---|---|---|---|---|---|
| | Dense | Sparse | Method | Features | | | |
| [22] | None | None | None | | Dependency Treelet SMT | - | 2006 |
| [23] | None | None | None | | None | - | 2012 |
| [24] | None | None | None | | Error Generation (EVP), POS Factored SMT, Error Selection | EVP | 2013 |
| [25] | None | None | None | | Hierarchical PB-SMT | | 2013 |
| [26] | None | None | None | | Treelet LM, Classifier | Gigaword | 2013 |
| [27] | Char-Level LD | None | Averaging | LM Inside SMT, Web-Scale LM | RB System, Error Generation (EVP), POS Factored SMT, Error Selection | EVP | 2014 |
| [28] | Word-Level L1 | Yes | None | | Error Selection | Wikipedia, Common Crawl | 2014 |
| [29] | None | None | None | | Classifier | Wikipedia | 2014 |
| [32] | None | None | None | | NNJM | Wikipedia | 2016 |
| [33] | Char-Level LD | None | SVM | LM, ALM, Length, SMT, Word Lexicon | None | - | 2016 |
| [34] | None | None | Perceptron | POS Tag, Shallow Parse Tag, Combined | None | Gigaword | 2016 |
| [35] | None | None | Averaging | SMT, Lexical and POS, LM, Contexts | None | Wikipedia | 2016 |
| [31] | None | None | None | | NNGLM, NNJM | Wikipedia | 2016 |
| [37] | LD, Edit Counts, OSM, WCLM | Yes | None | | None | Wikipedia, Common Crawl | 2016 |
| [38] | Edit Counts, OSM, WCLM | Yes | None | | Char-level SMT, NNJM | Common Crawl | 2017 |
| [39] | Word-level LD, Edit Counts, OSM, WCLM | Yes | None | | Char-level SMT, NMT Pipline | Common Crawl | 2018 |
| [36] | None | None | None | | Classifier, Speller | Wikipedia | 2016 |

A group of studies [33], [34], [35] paid more attention to n-best list reranking to solve the "first but not the best"

problem in the outputs of previous SMT based GEC systems. An extra rescoring post-processing also enabled more information to be incorporated into the SMT based methods, through which improvement based on the previous systems has been acquired. Empirical study [36] concerning distinguished approaches has compared the SMT and classification based approaches by performing error analysis of outputs, which also promoted a sound pipeline system using classification based error type-specific components, a context sensitive spelling correction system, punctuation and casing correction systems, and SMT. Also, system based on SMT only [37] described a baseline GEC system using task-specific features, better language models, and task specific tuning of the SMT system, which outperformed previous GEC models. Later research [38] further incorporated both the NNJM and a character-level SMT for spelling check into the baseline SMT based model and has shown the efficiency of their methods with an increase in the final performance. As the increasing development of the neural based methods in both the MT and GEC, the more effective interaction of those methods has been researched [39] based on their comparison between SMT and NMT, also hybrid models were proposed with higher testing score.

TABLE 3
NMT based GEC systems. The column "Training LM" is the same as that in Table 2.

| Source | Model | Framework | Input | Handling Misspelling | Training LM | Year |
|--------|-------|-----------|-------|---------------------|-------------|------|
| [45] | ED | RNN | token level | alignment, word-level translation | - | 2016 |
| [47] | | RNN | character level | character-level translation model | Common Crawl | 2016 |
| [49] | | RNN | token level, character level | character-level translation model | Common Crawl | 2017 |
| [68] | | RNN | token level | spellchecker | - | 2017 |
| [50] | | CNN | token level | spellchecker | - | 2018 |
| [67] | | RNN | token level | spellchecker | Wikipedia | 2018 |
| [39] | | RNN | token level | spellchecker | Common Crawl | 2018 |
| [53] | | Transformer | token level | spellchecker | Common Crawl | 2018 |
| [56] | | Transformer | token level | spell error correction system | Common Crawl | 2019 |
| [66] | | CNN | token level, sentence level | spellchecker | - | 2019 |
| [69] | PIE | Transformer | token level | spellchecker | Wikipedia | 2019 |
| [71] | ED | Transformer | token level | - | - | 2019 |

### 3.1.3 GEC-specific Features

Although the widely used SMT tools Moses [40] had contained dense and sparse features tuning towards BLEU, which was apparently under the machine translation setting, the direct exploitation of those methods seemed to be inappropriate [37]. Early in the CoNLL-2014 shared task, several methods [27],[28] which focused on the task-specific features have been proposed tailored the particularity of grammatical error correction. Systems using character-level Levenshtein distance as dense features [27], [33] and research investigating both specific dense and sparse features [28] all have demonstrated the improvement in the performance of SMT based approaches. Better interaction of dense features and sparse features were scrutinized in their later research [37], which cultivated a strong baseline SMT model. Here we survey the commonly used GEC-specific features that may offer an improvement to an SMT based model in GEC.

**Dense features.**
- Levenshtein distance. Both character-level Levenshtein distance and word-level Levenshtein have been used as dense features. Through the distance, the relation between target and source sentence could be modeled, especially the edit operations, which mainly reflect the correction patterns, are captured.

- Edit operation counts. Similar but a more refined and detailed version of Levenshtein distance feature is edit operation counts. Based on the Levenshtein distance matrix, the numbers of deletions, insertions, and substitutions that transform the source phrase into the target phrase are computed, noticeably the sum of these counts is equal to the original Levenshtein distance.

- Operation Sequence Model (OSM). Operation Sequence Model is introduced into Moses for machine translation [41]. These models are Markov translation models that in GEC setting can be interpreted as Markov edition models. Translations between identical words are matches, translations that have different words on source and target sides are substitutions; insertions and deletions are interpreted in the same way as for SMT.

- Word-class language model (WCLM). The injection of word-class information has shown their contribution in the machine translation task early from the IBM series of model. A more general used method used monolingual Wikipedia data to create a 9-gram word-class language model with 200 word classes produced by word2vec [42]. These features allow to capture possible long distance dependencies and semantical aspects in the SMT based model.

**Sparse features.** More fine-grained features can be extracted from the Levenshtein distance matrix as specific error correction operation types with or without context, by counting specific edits that are annotated with the source and target tokens that take part in the edit.

### 3.1.4 Web-scale Language Models
As demonstrated in equation 2 and 3, a target side language model $p(y)$ is already included in the noisy channel model or the more general log-linear model. However, the size of data when extracting the target side of training sets is so limited that additional web-scale language models are often trained on the large monolingual (local error-free) datasets. The monolingual corpus could be Wikipedia [29], [28], [32], [35], [31], [36], Gigaword [26], [34] and Common Crawl [28], [37], [38], [39]. Such LMs have been trained though tools such as KenLM or IRSTLM in a range of researches to further prompt the improvement of their systems' performance.

### 3.1.5 Neural Networks as Features

Although the translation model of SMT based approaches has shown the effectiveness by estimating phrase table, the discrete phrase table and the linear mapping on which translation probability based still limited the competent of model to generalize more to pattern beyond the training sets. Also, the global information was always ignored by the basic SMT based model. Although a range of language models and other sequence modeling features have been incorporated in several SMT based GEC models, context information concerning each word was lacked. Some works tackled these limitations using complementary neural networks as additional features, namely neural network global lexicon model (NNGLM) and neural network joint model (NNJM), which construct continuous representation space with non-linear mapping modeled [32], [31], [38].

**NNGLM.** A global lexicon model here is a feed forward neural network used to predict the presence of words in the corrected output by estimating the overall probability of hypothesis given the source sentence. The probability of a target hypothesis is computed using the following equation:

$$\log p(\hat{y}|x) = \sum_{i=1}^{T_{\hat{y}}} \log p(\hat{y}_i|x),  \quad (4)$$

where $p(\hat{y}_i|x)$ is the probability of the target word $\hat{y}_i$ given the source sentence $x$; $p(\hat{y}_i|x)$ is the output of the neural network.

**NNJM.** Joint models in translation augment the context information in language models with words from the source sentence. Unlike the global lexicon model, NNJM uses a fixed window from the source side and takes sequence information of words into consideration in order to estimate the probability of the target word. The probability of the hypothesis $h$ given the source sentence $x$ is estimated by the following equation:

$$\log p(\hat{y}|x) = \sum_{i=1}^{T_{\hat{y}}} \log p(\hat{y}_i|c_i),  \quad (5)$$

where $c_i$ is the context (history) for the target word $\hat{y}_i$. The context $c_i$ consists of a set of source words centralized by word $\hat{y}_i$ and certain number of words preceding $\hat{y}_i$ from the target sentence same as the language model does.

### 3.2 NMT Based Approaches

Although SMT based approach benefits from its ability to incorporate the large amount of parallel data and monolingual data as well as the auxiliary neural network components, it still suffers from the lack of contextual information and limited generalization ability. As a solution, many researches start to research NMT based approaches for GEC. With the increasing performances obtained by neural encoder-decoder models [43] in machine translation, neural encoder-decoder based models are adopted and modified. Compared to SMT based GEC systems, NMT based models have two advantages. First, neural encoder-decoder model learns the mappings from source to target directly from training parallel data, rather than the required features in SMT to capture the mapping regularities. Second, NMT based systems are able to correct unseen ungrammatical phrases and sentences more effectively than SMT based approaches, increasing the generalization ability [44].

In this section, we trace the development of representative works, which are the most commonly used as backbones combining with other techniques in GEC systems today. We leave the numerous techniques to Section 4, and focus only on works researching the design and training schema of neural models in GEC. As we will discuss, all the neural GEC systems are based on the encoder-decoder (ED) model, with an exception based on the parallel iterative edit (PIE) model.

### 3.2.1 Development of NMT Based Approaches

Yuan and Briscoe first applied NMT based models in GEC [45]. In this work, the encoder encodes the source sentence $x = \{x_1, x_2, ..., x_{T_x}\}$ as a vector $v$. Then, the vector is passed to the decoder to generate the correction $y$ through

$$p(y) = \prod_{t=1}^{T} p(y_t|y_{1,...,t-1}, v).  \quad (6)$$

At each time step, the target word is predicted with the vector and the previously generated words. Both the encoder and the decoder are RNNs composing of GRU or LSTM units.

Attention mechanism [46] is always applied to improve the output in each decoding step $i$ by selectively focusing on the most relevant context $c_i$ in the source. To be more specific, the hidden state in decoder is calculated by

$$s_i = f(s_{i-1}, y_{i-1}, c_i),  \quad (7)$$

where $s_{i-1}$ is the hidden state last step; $y_{i-1}$ is the word generated last step; $f(\cdot)$ is non-linear function. The variable $c_i$ is calculated as

$$c_i = \sum_{j=1}^{T_x} \alpha_{ij} h_j,  \quad (8)$$

where $h_j$ is the hidden state of encoder at step $j$ and $\alpha_{ij}$ is calculated as

$$\alpha_{ij} = \frac{e^{m_{ij}}}{\sum_{k=1}^{T_x} e^{m_{ik}}}.  \quad (9)$$

The variable $m_{ik}$ is a match score between $s_{i-1}$ and $h_k$.

Xie et al. firstly applied the first character-level neural encoder-decoder model for GEC [47]. Since operating at character-level increases the recurrent steps in RNN, a pyramid architecture [48] is applied to reduce computational complexity and the encoded representation of the input is obtained at the final hidden layer of the pyramid architecture. The decoder is attention-based and also has the pyramid architecture to reduce the calculation relating to attention mechanism.

Combining word-level NMT model and character-level NMT model, a hybrid NMT based GEC model with nested attention is proposed [49]. In this hybrid architecture, words that are in target vocabulary are generated by word-level decoder, while those who are out of target vocabulary are generated by character-level decoder. During each decoding step, the probability of each token is the product of probability of unknown words (UNKs) calculated by *softmax* function of word-level decoder $p$, and the probability of the character sequence in a token generated by the character-level decoder. To be more specific, the character-level decoder will select a source word $x_{z_s}$ according to

$$z_s = \arg\max_{k \in 0...T-1} \alpha_{sk}, \tag{10}$$

where $\alpha_{sk}$ is calculated by word-level attention. If the source word $x_{z_s}$ is in source word vocabulary, then the character-level decoder initializes the initial hidden state using $\hat{h}_t = ReLU(\hat{W}[c_s; h_s])$, where $h_s$ is the hidden state of word-level decoder. However, if the source word $x_{z_s}$ is out of source word vocabulary, then a nested attention is applied to make character-level decoder attend to $x_{z_s}$. By directly providing the decoder with access to the character sequence in the source word, out-of-vocabulary (OOV) problem could be better addressed in GEC.

Besides the widely applied RNN based NMT model [45], [47], [49], [39], the first CNN based NMT model on GEC became the first NMT based GEC systems outperforming SMT based GEC systems [50]. A multi-layer convolutional architecture [51] is designed and CNN based attention for sequence-to-sequence learning is applied. In each encoder layer, the source sentence is firstly embedded using word embedding and position embedding as $S \in \mathbb{R}^{h \times |S|}$, where $|S|$ is the number of tokens in the source sentence, and then linearly transformed into $H^0 \in \mathbb{R}^{h \times |S|}$ before it is fed into the first encoder layer. The output of the $l$th encoder layer is calculated as follow:

$$H^l = \text{GLU}(\text{Conv}(H^{l-1})) + H^{l-1}, \tag{11}$$

and the output of the final encoder layer HL is transformed into the output of encoder $E \in \mathbb{R}^{d \times |S|}$. During decoding, the embedding for the generated $n$ target words $T \in \mathbb{R}^{d \times n}$ is calculated the same as $S$, and is linearly transformed into $G^0 \in \mathbb{R}^{d \times n}$. The $l$th decoder layer computes an intermediate representation:

$$Y^l = \text{GLU}(\text{Conv}(G^{l-1})), \tag{12}$$

which is used for the calculation of attention:

$$Z^l = \text{Lin}(Y^l) + T, \tag{13}$$

$$X^l = (E + S) \times \text{softmax}(E^T \times Z^l), \tag{14}$$

$$C^l = \text{Lin}(X^l). \tag{15}$$

Then, the output of the $l$th decoder layer is calculated as follows:

$$G^l = Y^l + C^l + G^{l-1}. \tag{16}$$

The matrix $G^L$ is then linearly transformed to $D \in \mathbb{R}^{d \times n}$. The final column of $D$ is then mapped to the size of vocabulary to predict the $(n+1)$ word. The strength of CNN based models over RNN based ones is that CNN is more efficacious at capturing local context and thereby corrects a wider spectrum of grammatical errors. The long-term context information could be captured by the multi-layer architecture. Another advantage of CNN is that unlike RNN, where the number of non-linearity operation on source sentences increases linearly as the length of sentences, only a fixed amount of non-linearity operation would be conducted and thus more semantic information could be exploited.

With the proposal of Transformer [52], many NMT based GEC models replaced traditional RNN based encoder-decoder with Transformer [53], [54], [55], [56], [57], [58], [59], [58], [60], [61]. Transformer first encodes the source sentence into a hidden state through a stack of several identical blocks, each consisting of a multi-head self-attention layer and a forward layer. The multi-head self-attention is calculated as

$$Attention(Q, K, V) = \text{softmax}(\frac{QK^T}{\sqrt{d_k}})V, \tag{17}$$

where $Q, K, V$ represent the query matrix, key matrix and value matrix respectively and are calculated by linear transformation on input vectors. The variable $d_k$ is the dimension of column vectors in $Q$ and $K$. The decoder has the same architecture as encoder, but with an additional mutual attention layer over the hidden states. Compared to traditional RNN (with GRU units or LSTM units), one of the advantages of self-attention layer of Transformer is that it enables parallel computing so that the time spent on training GEC models could be shorten. Although Transformer reads the whole sentence at once without the order information, positional encoding could largely benefit the representation learning.

Copy mechanism was proposed to facilitate machine translation [62], and then was applied on GEC task and achieved state-of-the-art performance [56]. Copy mechanism is effective for GEC since only a few edits would be made on source sentences in GEC. In this work, the copy augmented model learns to copy unchanged source words into the target sentence directly. To be more specific, the final possibility of choosing the next word $\omega$ is composed of two parts, the possibility of generation softmax and the possibility of copying the word from source sentence

$$p(w) = (1 - \alpha) * p_t^{gen}(w) + \alpha * p_t^{copy}(w), \quad (18)$$

where $\alpha$ and $p_y^{copy}(\omega)$ are calculated with copy attention between encoder's output $H^{enc}$ and decoder's hidden state $H^{dec}$ t at step $t$:

$$p_t^{copy}(w|x, y_{1...t-1}) = \text{softmax}(K^T q_t), \quad (19)$$

$$\alpha = \text{sigmoid}(W \sum (A_t \cdot V)). \quad (20)$$

The matrix $K, V$ and the vector $q_t$ are calculated as

$$K, V, q_t = W_k H^{enc}, W_v H^{enc}, W_q h_t^{dec}, \quad (21)$$

and $A_t$ is given by

$$A_t = K^T q_t. \quad (22)$$

Integrating cross-sentence context is common among the researches about neural machine translation [63], [64], [65], but was firstly applied on GEC by Chollampatt et al. recently. In this work, they designed a cross-sentence convolutional encoder-decoder model with auxiliary encoder and gating, which was an extension on the previous work [50]. To be more specific, the auxiliary encoder encodes the previous two sentences into $\hat{S} \in \mathbb{R}^{d \times |\hat{S}|}$, and the corresponding output of the final encoder layer is $\hat{E} \in \mathbb{R}^{d \times |\hat{E}|}$. Auxiliary attention is involved and the auxiliary encoder representation at each decoder layer $\hat{C}^l$ is calculated the same as equation 15 but with different linear layers, $\hat{S}$ and $\hat{E}$. The output of lth decoder layer is now calculated according to:

$$G^l = Y^l + C^l + \Lambda^l \circ \hat{C}^l + G^{l-1}, \quad (23)$$

where $\circ$ is the element-wise production and $\Lambda^l$ is the gate calculated as follow [66]:

$$\Lambda^l = \sigma(\text{Lin}(Y^l) + \text{Lin}(C^l)). \quad (24)$$

Fluency boost learning aims at increasing the fluency and soundness of the source sentence to correct grammar errors [67]. The neural sequence-to-sequence model is trained with fluency boosting sentence pairs and multi-round inference strategies are applied to make full corrections to source sentences. Experiments have demonstrated the effectiveness of both the fluency boost learning and the inference strategies. We cover more details about the data generation methods and inference strategies in Section 5.2 and 4.4 separately.

Reinforcement learning has long been applied on other NLP tasks for post-processing, but was innovatively used on GEC to directly optimize parameters of neural encoder-decoder model [68]. The model (*agent*) searches the *policy* $p(h)$ directly according to:

$$\frac{\partial J(\theta)}{\partial \theta} = \alpha \mathbb{E}[\nabla \log(p(\hat{y}))\{r(\hat{y}, y) - b\}]. \quad (25)$$

The variable $b$ is baseline used to reduce the variance.

The *reward* $r(\hat{y}, y)$ is specified as GLEU. The parameters of the model are optimized using policy gradient algorithm.

Besides the direct sentence translation models discussed above, two models translating source sentences into edits that should be made to source tokens were successfully applied to GEC recently. The first is the PIE model [69], an improved version of the parallel model that has been previously explored in machine translation [70]. The PIE model is trained to label source tokens as edits from the edit space comprising of edit operations including copying, deletion, replacing with a 2-gram sequence, appending a 2-gram sequence and morphology transformation. The required supervision is collected by comparing source and target sentences in parallel training data. The PIE model is based on the architecture of BERT, but with additional inputs and a self-attention mechanism to better predict replacing and appending. Unlike the commonly used ED model, the PIE model generates the output concurrently, much faster than the encoder-decoder model. The second is the LaserTagger [71] that generates edit tags for source sentences. The LaserTagger is composed of a BERT encoder with 12 self-attention layers and an autoregressive single layer Transformer decoder. The edit tags include all the possible combination of two base tags: KEEP and DELETE, and many phrases that should be inserted before the source tokens. The phrases are the 500 most frequently n-grams that are not aligned between source sentences and target sentences in training data. An algorithm is designed to convert the training examples into corresponding tag sequences. The very advantage of generating edits is that the edit space is rather small than the target vocabulary space.

### 3.2.2 Handling Misspellings

Due to the limitation of vocabulary size in training NMT models, it is unrealistic to cover all the misspelled words in the vocabulary. So, several methods are applied to correct misspelling errors in NMT based GEC systems.

- **Alignment+Translation.** The misspelled words can be viewed as UNKs, and the problem could be solved by first aligning the UNKs in the target sentence to their corresponding words in the source sentence with an aligner and then training a word-level translation model to translate those words in a postprocessing step [45].

- **Character-level translation model.** As discussed before, some NMT based models are character-level, so the problem of misspelling errors is naturally solved [47], [49].

- **Spellchecker.** Many NMT based GEC systems utilize a spellchecker[1] or open-source spellchecking library[2] in preprocessing [68], [50], [53], [67], [66], [39].

---

[1] https://azure.microsoft.com/en-us/services/cognitiveservices/spellcheck/

[2] https://github.com/AbiWord/enchant

TABLE 4
Classification based GEC systems.

| Source | Target Error Type | Features | | Year |
|--------|-------------------|----------|---|------|
| [72] | article | Linguistic Feature Engineering | | 2006 |
| [73] | preposition | | | 2007 |
| [75] | | | | 2008 |
| [76] | | | | 2010 |
| [74] | | | | 2008 |
| [77] | article, preposition | | | 2008 |
| [78] | | | | 2010 |
| [79] | | | | 2010 |
| [80] | article, preposition, verb form, none number, sub-verb agreement | | | 2013 |
| [81] | article, preposition, verb form, none number, sub-verb agreement word form, orthography and punctuation, style | | | 2014 |
| [82] | article | Deep Learning | CNN | 2015 |
| [83] | article, preposition, verb form, none number, sub-verb agreement | | biRNN | 2017 |
| [86] | | | | 2018 |
| [84] | article, preposition, verb form, noun number, sub-verb agreement, comma | | | 2019 |
| [87] | all | | | 2019 |

- **Spell error correction system.** Zhao et al. built a statistical-based spell error correction system, and used it for correction of all the misspelling errors in their training data [56].

We recapitulate the GEC systems with NMT based approaches that we have introduced in Table 3. Except for the final performance achieved by the combination of basic approaches and other performance boosting techniques, we also include the experiment results brought by the NMT approaches themselves for *direct* comparison. We describe the performance boosting techniques in Section 4.

### 3.2.3 Incorporating Language Models

Similar to SMT based approaches, many NMT based approaches train n-gram language models on large monolingual data and integrated them into the beam search decoding phase to rank the hypothesises with consideration of fluency. To be more specific, a hypothesis's score $s(\hat{y})$ at the decoding step $k$ is calculated as follows:

$$s(\hat{y}_{1:k}|x) = \log p_{NN}(\hat{y}_{1:k}|x) + \lambda \log p_{LM}(\hat{y}_{1:k}), \quad (26)$$

where $p_{NN}$ is given by neural translation model and $p_{LM}$ is given by the n-gram language model weighted by $\lambda$. The monolingual corpus could be Common Crawl [47], [49], [39], [53], [56] and Wikipedia [67].

### 3.3 Classification Based Approaches

In classification based approaches, a multi-class classifier is trained to predict the correct word in the confusion set, based on features of the word's context from artificial feature engineering or deep learning models. Given the text to correct, for each token in the text occurring in the confusion set, the classifier predicts the most likely candidate. The classifier is trained on native or non-native speaker data, thus alleviating the requirement on annotated data, which is advantageous compared to supervised methods. The commonly adopted sorting algorithms including Maximum Entropy (ME), Naive Bayes, Decision Tree and Averaged Perceptron. Although classification approaches once were popular, they are not commonly adopted today, so we survey some typical works in general.

Most of the prior classification based approaches focus on the utilization and improvement of traditional elaborated feature engineering. During this period, approaches mainly focus on identifying or correcting article errors and preposition errors. Han and Leacock trained a maximum entropy classifier to select articles for noun phrases based on 11 features, most of which combines lexical and syntactic information, for example, the head word and its part of speech (POS) tag [72]. Chodorow et al. similarly trained a maximum entropy classifier to detect preposition errors by predicting the most probable candidate among 34 prepositions based on 25 contextual features, for example, the preceding and following word and POS [73]. Based on the previous work, De Felice and Pulman selected some new contextual features, and achieve an accuracy of 70.06% and 92.15% respectively [74]. Besides, the features for training and a series of filters and threshold are combined with ME approach so that the classifier output can be constrained [75]. Continuously, Tetreault et al. successfully added 14 parse features to baseline preposition model, and results showed statistically significant increased accuracy on native speaker test data [76]. Gamon et al. trained decision tree classifiers based on features including relative position, string and POS of tokens. For article errors and position errors, 2 separate classifiers are trained respectively. One for deciding whether or not a determiner/ preposition should be present and other predicting the most likely choice [77]. Gamon combined classifiers with language models and proposed metaclassifier in correcting articles and prepositions. The metaclassifier is trained using the features transforming from the output of the traditional classifiers and language models, which score the suggested correction. Slightly different, the meta-classifier is trained on error-annotated data. Each error annotation provides the information that whether the suggested correction is correct or incorrect [78]. Instead treating all prepositions equally when training classifiers, the confusion set could be restricted on specific several candidates which are frequently observed in occasions of misuse in non-native data, which is called L1-dependent confusion set, since the distribution of preposition errors differ by L1. Two methods that train classifiers based on L1dependent confusion set are proposed. In the first methods, only prepositions in L1-dependent confusion set are viewed as negative examples for each source prepositions. In the second methods, the all-class multi-classifier is trained on augmented data taking account into the L1 error statistic. The restriction increases both the precision and recall compared to previous work [79].

Later, some researches extended previous methods so that the classification based approaches can be applied on correcting more types of errors, instead of article errors and preposition errors only. The University of Illinois System trained classifiers to correct noun number errors, subject agreement errors and verb form errors. The confusion sets are composed of morphological variants of the source word, such as noun plurals, gerund and past tense of verb. This system ranked first in CoNLL-2013 shared task [80]. Continuously, in CoNLL-2014 shared task, the Illinois Columbia System extended by incorporating another three error-specific classifiers: word form errors, orthography and punctuation errors and style errors. They also applied joint inference to address inconsistent corrections suggested by multiple classifiers [81].

More recently, deep learning was combined with classification based system. Without the reliance on traditional elaborated feature engineering, Sun et al. firstly employed CNN to represent the context of source articles and predict the correct labels [82]. Wang presented a classification based study which trained deep GRU to represent context and predict the correct word directly. For the source word $\omega_i$ in the sentence, the context vector is calculated as follow:

$$biGRU(w_{1:n}, i) = lGRU(w_{1:i-1}) \oplus rGRU(w_{i+1:n}), \quad (27)$$

Where lGRU and rGRU read the words from left to right and GRU reading the words from right to left in a given context respectively. Then, the target word t is predicted according to:

$$t = \text{softmax}(ReLU(biGRU(w_{1:n}, i))). \quad (28)$$

For each error types involved (article, preposition, verb form, subject agreement, none number), a deep bidirectional GRU was trained [83]. Following the research above, Kaili et al. proposed two attention mechanism to predict correct words better. The first attention considers context words only, while the second attention take into account of the source word. The second attention is used for correction of none number errors and word form errors, since the interaction between the source word and its context is important to correction of these two types of errors. Li et al. trained bidirectional GRU to correct errors including subject-verb agreement, article, plural or singular noun, verb form, preposition substitution, missing comma and period comma substitution [84]. They also trained a pointer context model [85] to correct word form error [86]. Makarenkov et al. designed a bidirectional LSTM to assign a distribution over the vocabulary where correction tokens are selected and to suggest proper word substitution. A postprocess based on POS tagger is appended to filter out the suggested words with different property from the source tokens [87].

We summarize the introduced classification based GEC systems in Table 4, with their target error types and where the features come from.

### 3.4 LM Based Approaches

The SMT based methods and NMT based methods are all supervised, thus require large amounts of parallel training data. However, unsupervised approaches based on LM are applied to GEC and achieve comparable performance with supervised methods. The very advantage of LM based approaches is that they do not need large parallel annotated data, given the fact that large amounts of parallel data is not available in GEC. As a result, LM based GEC systems are always developed for low-resource situation. Most of LM based GEC methods use a LM to assess the hypothesis for correction, where hypothesis is generated by making changes to source ungrammatical sentence according to the designed rules or by substituting tokens in source sentence with words selected from *confusion sets*. In this section, we first describe the basic and the optimized LM based approaches, then give more details about how to generate confusion sets.

N-gram language model can be used to compute a feature score for each hypothesis as follow:

$$score_{LM} = \frac{1}{T_{\hat{y}}} \log p_{LM}(\hat{y}). \quad (29)$$

Hypothesis with the highest score would be added to the search beam and then modified to generate another set of hypotheses. This iteration does not end until the beam is empty or the number of iteration has achieved a threshold [88]. Following this, Bryant and Briscoe re-evaluated LM based GEC on several benchmark datasets. Language model is used to calculate the normalised log probability of the hypothesis. The process of generating confusion set and evaluating hypothesis is also iterated following previous work. Without annotated data, this work achieves comparable performance with state-of-the-art systems testing on JFLEG test set. This is because the annotation of JFLEG is fluency oriented, thus making language model based approaches is more powerful [89].

A problem of neural LM based GEC is that the space of possible hypothesis is rather vast. To solve this issue, finite state transducer is used to represent large structured search space, constraining the hypothesis space while also promising that it is large enough to involve admirable corrections [90]. Several transducers are designed including input lattice $\hat{I}$ which maps the input sentence to itself, edit flower transducer $\hat{E}$ which makes substitution with cost, penalization transducer $\hat{P}$, 5-gram count based language model transducer $\hat{L}$ and transducer $\hat{T}$ which maps words to byte pair encoding (B). $\hat{P}$ and $\hat{L}$ are used to score the hypothesis space. The best hypothesis is searched using the score of the combined FST and a neural language model. Additionally, if large annotated data is available to train SMT and NMT models, the n-best list of SMT system is used to construct $\hat{I}$ and NMT score is incorporated into the final decoding procedure. Based on this, Stahlberg and Byrne composed input transducer $\hat{I}$ with a deletion transducer $\hat{D}$ to delete tokens that are frequently deleted. They also constructed a insertion transducer $\hat{A}$ that restrict insertion to three specific tokens ",", "-" and "'s". Their LM based approaches achieved higher score on CoNLL-2014 test set [91].

Confusion set is an important element in LM based GEC, since language model selects the most possible hypothesis from it. Dahlmeier and Ng generated hypothesis by (1) replacing misspelled word with correction; (2) changing observed article before Noun Phrase; (3) changing the observed preposition; (4) inserting punctuation; (5) altering noun's number form (singular of plural) [88]. Similarly, Bryant and Briscoe created confusion set by (1) applying spell checker CyHunspell on misspelled words; (2) altering morphological forms of words using automatically generated inflection database; (3) changing the observed article and preposition [89]. Besides, WikEd Error Corpus was also used to create a part of the confusion set [92].

Flachs et al. combined pre-trained language models such as BERT and GPT-2 with noisy channel model for GEC. For each word $w$ in a given sentence, the noisy channel model estimates the probability that w is transformed from a candidate $c$ in the confusion set. The idea behind this approach is that for any given word $w$, there is a genuine word $c$ that passes through the noisy channel and transforms into $w$. The goal is to choose the most possible genuine word $c^*$:

$$c^* = \arg\max_{c \in C} P(c|w). \qquad (30)$$

According to Bayes' principle, equation could be written as

$$c^* = \arg\max_{c \in C} P(w|c) * P(c), \qquad (31)$$

where $P(w|c)$ is estimated by noisy channel model; $P(c)$ is given by pre-trained language model (BERT, GPT-2) and $C$ is the generated confusion set for $w$ [92].

### 3.5 Hybrid Approaches

Apart from models we have discussed, a range of researches also integrated several models into their systems to seize profit from different models in order to obtain better performance. Such systems with sub-models will be surveyed in this section as hybrid methods, which often contain heterogeneous sub-components. It is worth noticing that ensemble methods (section 4.3) also involve content concerning model combination which is similar to this section. However, in this survey we restrict the ensemble methods to the techniques assistant better combination of outputs produced by a range of models. The major difference between hybrid and ensemble in this survey is that the hybrid methods integrate several sub-system to construct a complete correction process, whereas ensemble methods use several independent whole system, often combined at the output level, to improve the correction performance.

Yoshimoto et al. and Felice et al. employed different sub-components in their system to gear with specific error types in the source sentence and outputs from sub-systems, which are partially corrected, are merged to produce an error-free sentence. More specifically, Yoshimoto et al. used three systems to deal with all five error types in the CoNLL2013 shared task, including a system based on the Treelet language model for verb form and subject verb agreement errors, a classifier trained on both learner and native corpora for noun number errors and an SMT based model for preposition and determiner errors[26]. Felice et al. combined both a rule based system and an SMT based system. Their outputs are combined to produce all possible candidates without overlapping correction and then a language model is trained to rerank those candidates [27].

Pipeline methods are also researched in combining subsystems, where the output corrected by one system is passed as an input to a following correcting system. Rozovskaya and Roth first applied a classification model followed with an SMT based system, since the SMT-system owns the ability to handle with more complex situation than classifiers [36]. A similar framework is proposed by combining the more powerful neural classifiers and SMT based system [86]. Grundkiewicz and Junczys-Dowmunt constructed an SMT-NMT pipeline and experiments in the research have shown complementary corrections have been made. They also explore using the NMT system to rescore the n-best hypothesis obtained through SMT system to improve fluency of final outputs [39].

### 3.6 Others

Some prior works used hand-crafted rules to correct some specific types of grammar errors. Rules are designed for correcting the forms of those verbs related to prepositions and modal verbs [93] and noun errors based on a dictionary composed of 250 most common uncountable nouns ("time" is an exception and it has its own rule) [94]. Besides, syntactic n-grams are combined with the rule based system, as a result of which whenever the system finds preposition, it would search the list of extracted patterns and match correctly [94].

<div align="center">

IV. PERFORMANCE BOOSTING TECHNIQUES

</div>

While various approaches are successfully applied to solve GEC problem, a wide range of performance boosting techniques beyond basic approaches are also created and developed in order to facilitate the GEC systems to achieve better performance. In this section, we divide the typical techniques into several branches and describe how they are applied in detail. We intentionally describe the techniques in this section and the basic approaches (section 3) in separation in order to distinguish the improvement brought by techniques, although most techniques are combined with NMT based GEC approaches. In this way, basic GEC models could incorporate multiple techniques for more powerful GEC systems. Data augmentation strategies can be also viewed as performance boosting techniques, however, due to the large amount of researches and the broad application of them, we describe data augmentation methods in Section 5 individually.

It is worthy to mention that some training techniques including dropping out word embeddings of source words [95], checkpoint average [96], byte pair encoding (BPE) [97] and domain adaptation [53] are commonly used in training GEC systems. However, these tricks are not initially proposed or adapted to improve the performance of GEC systems. Since numerous performance boosting techniques are applied and combined with each other, it is unrealistic to give all of them a detailed description. So, we pay attention to techniques that are more specific to GEC, especially researches that are intended to conduct methodological exploration.

### 4.1 Pre-training

Sequence transfer learning has shown huge improvements on the performance of many NLP tasks, since a pre-trained model often obtained more reasonable initial parameters than using random initialization. Such pre-trained models can improve the convergence speed and training efficiency, leveraging knowledge acquired from related data while reducing the need for high-quality data for tasks where sufficient corpus are not available. Since GEC is always treated as low-resource machine translation task [53], various pretraining approaches are applied on GEC to improve the performance of models. The main discrepancies among these pre-training approaches are (1) whether the pre-training uses GEC related data and (2) what part of the parameters of the whole model is initialised with pre-trained model.

#### 4.1.1 Pre-training with Artificial Parallel Data

Due to the lack of high-quality data, many studies on GEC focused on the efficient usage of pseudo parallel training data, most defining a pre-training task to introduce the information of error patterns comprised in GEC artificial parallel data. An NMT based GEC system typically adopted this method to pre-train its neural seq2seq model as a denoising autoencoder (DAE) [98]. Given the original sentence y and its noised counterpart $\hat{y}$, the training of autoencoder aims at minimizing the distance between $y$ and $\hat{y}$. During the pre-training, the model learns to reconstruct the input sentence and thus is capable of making corrections. The parameters of both encoder and decoder are pre-trained.

The pre-trained model could be used for the following training procedure with two strategies: *re-training* and *fine-tuning*. The main difference between re-training and finetuning is that the learning rate and parameters of optimizer are reset in re-training strategy, while they are maintained in fine-tuning scenario, although some work may fine-tune the pre-trained model with a smaller learning rate.

Many Transformer based GEC systems adopted pre-training on artificial parallel GEC data [57], [56], [55], [99], [100]. We cover more details about artificial data generation in Section 5. Besides the synthetic parallel data, *Wikipedia* revision history could also be a source for pre-training [54].

#### 4.1.2 Pre-training with Large Monolingual Data

Pre-training on large error-free monolingual data can also boost the final performance. In neural encoder-decoder models, pre-training the decoder as a language model on large monolingual could largely benefit the final performance, since the architecture of decoder of sequence learning model is the same as language model [101]. Unlike the pre-training method described in last section, only the parameters of decoder are pre-trained, and the parameters of encoder and attention mechanism are moved away during pre-training. Many NMT based GEC systems benefit from this technique with a few discrepancies in the source of monolingual data [53], [56], [102], [84].

Besides the NMT based systems, the parallel edit model [69] can also be pre-trained using large amounts of error-free sentences in a way much like the training of BERT, predicting the arbitrary masked token combining both forward context and backward context.

## 4.2 Reranking

Compared to other performance boosting techniques which are often integrated inside the models, reranking, as a kind of post-processing, is a more individual component employed after the whole correction process. Often n-best outputs (as candidates or hypothesis) with the highest probability given by the correction model are rescored during the reranking and the optimal candidate will be selected according to the new scores as the final output. The aim of n-best list reranking, in GEC, is to rerank the correction candidates produced by the previous components using a rich set of features that are not well-covered before, so that better candidates can be selected as "optimal" corrections. By appending a reranking component, (1) linguistic information and features tailored to GEC could readily be introduced into correction system; (2) outputs from different grammatical error correction systems can be incorporated; (3) some global features could be exploited without the decoding processing being modified. Empirical studies showed the deficiency of systems that the best hypothesis may not be the optimal correction [34] and a range of researches have employed reranking demonstrating that there is considerable room for improvement brought by the reranking component. We summarize the commonly adopted features and how to incorporate them respectively.

### 4.2.1 Features

- **Language model.** An n-gram language model is trained on large monolingual corpora to give a score $p_{LM}(\hat{y})$ to each hypothesis. The score is calculated by summing all the n-gram log possibilities together and then normalize it by the length of hypothesis. Many systems train a 5-gram language model [50], [103], [61]. Besides, the *masked language model* probabilities computed by BERT can also be used for reranking [66]. For each token $\hat{y}_i$ in the hypothesis $\hat{y}$, they replace $\hat{y}_i$ with [MASK] token, and calculate the probability of the masked token. This feature score is calculated as

$$f_{BERT}(\hat{y}) = \sum_{i=1}^{T_{\hat{y}}} \log p_{BERT}(\hat{y}_i|\hat{y}_{-i}). \quad (32)$$

- **Sentence-level correctness.** A neural sequence error detection model is trained to rerank the n-best hypothesizes output by MT based model. The model assigns a probability $p(\hat{y}_i)$ to each token indicating the likelihood that the token is correct. Given a candidate in the n-best list, the probability of each token being correct is $\Sigma_{i=1}^{T_{\hat{y}}} \log p(\hat{y}_i)$. The neural error detection model can be the combination of two bidirectional LSTM-RNNs to encode both the character and the context of the token [104], [59] or trained with auxiliary context predicting tasks [104],

where the loss function is modified into the following:

$$E = -\sum_{t=1}^{T} \log p(y_t|x_t...x_T) \quad (33)$$
$$-\gamma \sum_{t=1}^{T-1} \log p(x_{t+1}|x_1...x_t)$$
$$-\gamma \sum_{t=2}^{T} \log p(x_{t-1}|x_t...x_T).$$

The sentence-level correctness score could also be predicted by BERT, which is fine-tuned on learner corpus [61].

- **Edit operation.** Three features relating edit operation is always combined with other features, including the numbers of token-level substitutions, deletions and insertions between source and hypothesis [50], [61]. This feature could be replaced with a similar Levenshtein Distance feature [104].

- **Right-to-left model.** Inspired by the application of right-to-left model in machine translation [97], some GEC systems rerank hypothesis using the scores of right-to-left model to better incorporate the context of each word [57], [99].

- **Neural quality estimation** Chollampatt and Ng proposed the first *neural quality estimation* model for GEC and used the quality estimation score as a feature in reranking, which yields statistically significant improvement on base GEC model. In this work, the neural quality estimation model is composed of a predictor and an estimator. The predictor is an attention-based multi-layer CNN model, trained on parallel source and target sentences, to predict the probability of the tokens in target sentence given the source sentence and the context of target token. The estimator is also CNN based, trained on source sentences and system hypothesises, predicting the quality estimation score of hypothesises. The golden standard score given by GEC evaluation metric $M^2$ is used as label in the training [102].

- **Syntactic Features.** Several global syntactic features could be injected into reranking process. A range of researches use lexical features that the words occurring in the source side and target side of an edit and their parts-of-speech (POS) as features. The lexical features can determine the choice and order of words and the POS features can determine the grammatical roles of words in the edit within a hypothesis [33],[35]. Other sequential syntactical features extract from structures like dependency parse tree were also researched [34].

- **Translation Model Score.** Reusing the score from translation model can partially preserve the correction information in the reranking process. Such features extracted from the correction models like decoding scores and n-best list ranking information (represented linearly or non-linearly) could be incorporated for rescoring [33].

- **Adaptive language model.** Adaptive LM scores are calculated from the n-best list's n-gram probabilities. N-gram counts are collected using the entries in the n-best list for each source sentence. N-grams occurring more often than others in the n-best list get higher scores, ameliorating incorrect lexical choices and word order. The n-gram probability for a hypothesis word $\hat{y}_i$ given its history $\hat{y}_{i-n+1}^{i-1}$ is defined as:

$$p_{n-best}(\hat{y}_i|\hat{y}_{i-n+1}^{i-1}) = \frac{count_{n-best}(\hat{y}_i, \hat{y}_{i-n+1}^{i-1})}{count_{n-best}(\hat{y}_{i-n+1}^{i-1})}. \tag{3}$$

The sentence score for the $s$-th candidate $\hat{y}^s$ is calculated as:

$$score(\hat{y}^s) = \frac{1}{T_{\hat{y}}} \log(\prod p_{n-best}(\hat{y}_i|\hat{y}_{i-n+1}^{i-1})), \tag{3}$$

where it is normalized by the sentence length [33].

- **Length feature set.** Length features are often used to penalize overhaul in correction process. Unnecessary deletions or insertion will be limited by introducing those length ratios:

$$score(\hat{y}^s, s) = \frac{N(\hat{y}^s)}{N(s)}, \tag{36}$$

$$score(\hat{y}^s, \hat{y}^1) = \frac{N(\hat{y}^s)}{N(\hat{y}^1)}, \tag{37}$$

$$score(\hat{y}^s, \hat{y}^{max}) = \frac{N(\hat{y}^s)}{N(\hat{y}^{max})}, \tag{38}$$

$$score(\hat{y}^s, \hat{y}^{min}) = \frac{N(\hat{y}^s)}{N(\hat{y}^{min})}, \tag{39}$$

where $\hat{y}^s$ is the s-th candidate; $\hat{y}^l$ is the 1-best candidate (the candidate ranked 1st by the correction system); $N(\cdot)$ is the function calculating the sentence's length, thus $N(\hat{y}^{max})$ returns the maximum candidate length in the n-best list for that source sentence and $N(\hat{y}^{min})$ returns the minimum candidate length [33].

### 4.2.2 Incorporation

The most commonly used incorporation method is calculating the weighted sum of the scores given by multiple feature functions and selecting the hypothesis with the maximum total score, as expressed in the following equation:

$$\hat{y}^* = \arg\max_{\hat{y}} \sum_{i=1}^{m} \lambda_i f_i(\hat{y}), \tag{40}$$

where $m$ is the number of feature functions; $\lambda_i$ is the weight for the $i$th feature; $f_i(\cdot)$ is the feature function; $\hat{y}$ is the hypothesis being reranked.

Also, several studies trained a discriminative ranking model to score the n-best list derived from the preset features extracted from those hypothesises [33], [34]. Such incorporation of the features closely correlates their ranking performance with the ranking model they train. Weighting feature functions differently as the parameter tuning does in the previous methods, the ranking models serve the same purpose of obtaining a learnable model to better consider different feature functions' score according to the performance.

### 4.3 Model Ensemble

Although model ensemble does not improve the performance of individual model or algorithm in essence, combining different systems has the potential to improve both recall and precision. Recall could be increased when systems focusing different aspects of corrections are well-integrated so that errors could be identified much more comprehensively. Precision could be increased by utilizing the fact that correction produced by multiple systems will give us more confidence about its correctness. Also, we should notice that, besides GEC, various ensemble methods have been proposed and exploited over different tasks in NLP. A range of more sophisticated methods have been successfully used in fields such as Named Entity Recognition (NER) and Entity linking (EL). In this section, we investigate a common way to combine models, which achieves the ensemble during decoding, and several empirical studies introducing more complex ensemble methods tailored to GEC.

### 4.3.1 Better Search in Decoding

A widely used ensemble method utilizes different systems when searching the n-best list during decoding. Traditionally, a range of models (homogeneous or heterogeneous) are trained and fine-tuned under a distinct settings separately and then combined during the decoding process, where the prediction score of each model is averaged. Model ensemble is widely used in the development of GEC systems together with beam search to obtain a more reliable n-best list. For example, an ensemble of 4 Transformer based machine translationmodelsbringshigherscoresontheBEA-2019test set [58], [91]. Similar systems also made use of an ensemble of 4 multi-layer CNN models [50]. More ensemble systems were explored of 3 sets neural GEC models, each set consisting 4 identical multi-layer CNN, while disparate training strategies and techniques, including label smoothing and source word dropout, are applied on training the three sets of models [102]. Since there are so many systems based on this simple ensemble pattern, and it is not restricted in GEC, we cover no more information about instances and details.

### 4.3.2 Recombining the Edits

A large proportion of GEC systems regard the correction task as a special machine translation problem, which maps the original sentence and corrected sentence to the source language sentence and target language sentence in the MT respectively. However, a significant difference between GEC and MT should be noticed that commonly the output of an MT system is produced and evaluated as a whole, whereas an output derived from GEC systems could be seen as a combination of edits or corrections which can be evaluated separately. The pilot implementation of this idea was done by a system taking advantage of outputs from both a classification and an SMT approaches [29]. In this work, pairwise alignments are first constructed using MEMT forming a confusion network and then the one-best hypothesis is produced during a beam search progress. Similar to MEMT, they employ features including language model, matches, backoff and length features to rescore both partial and full hypothesis. Their methods allow switching among all component systems, flexibly combining their outputs.

More recently, there have been some successful GEC systems that use more sophisticated ensemble strategies to combine multiple corrections extracted from outputs produced by different systems based on the fact aforementioned. Li et al. presented a ensemble method to integrate the output of 4 CNN based ensemble GEC models and 8 Transformer based GEC models and solved possible conflict. They first built up a confidence table for each individual system, consisted of the precision and $F_{0.5}$ of each error type given by ERRANT toolkit. Then, they designed three rules to merge the corrections output by different systems using the precision as the confidence of each correction, increasing the confidence of identical corrections while discarding the conflicting correction with lower confidence. Three types of model ensemble with different combination of CNN based and Transformer based model are investigated [84]. Kantor et al. proposed a ensemble method to combine different GEC systems at a higher scale, treating each individual system as a black-box. Their method merged multiple M2 format files of disparate GEC systems by splitting the M2 format files and identifying the probability that an edit of an error type is observed in a subset of edits. An optimization problem should be solved to determine the optimal probability for each error type [105].

### 4.4 Iterative Correction

Iterative correction in GEC aims at correcting source sentences not in single round decoding as traditional sequence generation, but instead in multiple round inference. The generated hypothesis is not be regarded as final output, but is fed into the model to be edited in following iteration. This is because some sentences with multiple grammar errors may not be corrected in only one decoding round, which requires higher reference ability of the model. The inspiration is from iterative beam search decoder [106], which makes correction on ungrammatical sentence through multiple rounds. In each round, the decoder searches the hypothesis space to select the best correction for the source sentence. The selected sentence is input to decoder as source sentence in the next iteration and is incrementally corrected. Some language model based approaches rely on this algorithm [88], [89].

Fluency oriented iterative refinement [67] corrects source sentence through multiple round inference until the fluency of hypothesis does not increase. The fluency is defined as follows:

$$f(x) = \frac{1}{1 + H(x)}, \tag{41}$$

$$H(x) = -\frac{\sum_{i=1}^{T_x} \log p(x_i | x_{<i})}{T_x}. \tag{42}$$

The variable $p(x_i | x_{<i})$ is given by a language model.

Some works also proposed an iterative decoding algorithm and successfully applied it to the model trained on out-of-domain Wikipedia data. In each iteration, an identity cost and every non-identity cost of hypothesis in the beam are calculated, and the hypothesis with minimal non-identity cost is maintained. If the minimal nonidentity cost is less than the product of an identity cost and a predetermined threshold, this hypothesis is viewed as a correction and input for next iteration. The model continues to correct the input sentence until no more edit is required [54], [60], [107].

Other tricks which could be viewed as iterative correction includes feeding the output into the ensemble translation model for a second pass correction [84], letting the SMT based model and classifiers take turns to make correction until no more alter is required [86], and parallel iterative edit model [69] which takes the output of the model as the input for further refinement until the output is identical to a previous hypothesis or the iteration number achieves the maximum round.

### 4.5 Auxiliary Task Learning

Although auxiliary task learning has been proved to be beneficial on many NLP tasks, not so many GEC works involve this technique. The most common auxiliary task for GEC is grammar error detection (GED), both word-level and sentence-level.

**Word-level GED.** Word-level GED could be treated as a token-level labeling task [56]. In particular, the token-level labeling task assigns a label to each source token indicating whether the source token is right or wrong, based on hidden representation of the source token $h_i^{src}$ :

$$p(label_i | x) = \text{softmax}(W_t h_i^{src}). \tag{43}$$

The model can learn more about the source token's correctness with this auxiliary task.

**Sentence-level GED.** Sentence-level GED could be treated as a sentence-level labeling task [59]. The model is trained to predict whether source sentences are grammatical or not based on hidden representation of the source sentence $H^{src}$:

$$p(label|x) = \text{softmax}(W_s H^{src}). \quad (44)$$

This classification task could be modified into sentence-level copying task, which enables the model to trust more about the source sentences when they are grammatical [56]. Besides, the sentence-level GED task could also be combined with sentence proficiency level prediction task [108]. A BERT is trained to predict both the binary GED label (whether correction is required) and the 3 proficiency level label simultaneously, and fine-tuned on each set of sentences with the same proficiency level.

### 4.6 Edit-weighted Objective

Another performance boosting technique is to directly modify the loss function in order to increase the weight of edited tokens between source and target [53], [102], [58], [60], [56]. To be more specific, suppose that each target token $y_j$ could be aligned to a corresponding source token $x_i : a_t \in \{0, 1, \dots, T_x\}$. If $y_j$ differs from $x_i$, then the loss for $y_j$ is enhanced by a factor $\Lambda$. The modified loss is as follows:

$$L(x, y, a) = -\sum_{t=1}^{T_y} \lambda(x_{a_t}, y_t) \log p(y_t | x, y_{<t}), \quad (45)$$

$$\lambda(x_{a_t}, y_t) = \begin{cases} \Lambda & x_{a_t} \neq y_t \\ 1 & otherwise \end{cases}. \quad (46)$$

## V. Data Augmentation

Data augmentation has always been explored in GEC, which is known as Artificial Error Generation (AEG), since supervised models suffer from the lack of parallel data and low quality. Unlike machine translation, large magnitude of parallel training data in GEC is not currently available. Thus, a wide spectrum of data augmentation methods incorporating pseudo training data were studied and applied in order to ameliorate the problem. Most importantly, data augmentation methods should capture commonly observed grammatical errors and imitate them when generating pseudo data. In this section, we survey some typical data augmentation methods, most of which could be divided into two kinds: *noise injection* and *back translation*.

It is worthy to note that the generated data can be typically used in two approaches. First, append the generated data to existing training data and use the combined data for training. In this approach, the pseudo data is viewed equivalent as real training data. Second, use the generated data for pre-training the neural model, and then use the real training data for fine-tuning. We have discussed pretraining in Section 4.1. It is demonstrated that when used for *pre-training*, larger amounts of pseudo data yield better performance, while

*equivalent* setting does not show significant improvement in final result [99]. As a result, many neural based GEC methods adopt pre-training to incorporate a large magnitude of artificial data.

### 5.1 Noise Injection

#### 5.1.1 Deterministic Methods

Among the data augmentation methods that corrupt error-free data by adding noise or applying noising function, some methods directly inject noise according to predefined rules, while others inject noise with consideration of more linguistic features. We first summarize direct noise injection methods, which are also called *deterministic* methods.

Izumi et al. firstly used predefined rules to create a corpus for grammar error detection by replacing preposition in original sentences with alternatives [109]. Brockett et al. used predefined rules to alter quantifiers, generate plurals and insert redundant determiners [22]. Lee and Seneff defined rules to change verb forms to create an artificial corpus [110]. Ehsan and Faili defined error templates to inject artificial errors to treebank sentences when original sentences matched the error templates [111]. More recently, similar to direct noising methods applied in low-resource machine translation [112], Zhao et al. generated artificial data using corruption including (1) insertion of a random token with a probability (2) deletion of a random token with a probability (3) replacing a token with another token randomly picked from the dictionary probabilistically (4) shuffling the tokens [56]. Lichtarge et al. applied character-level deletion, insertion, replacement, and transposition to create misspelling errors [54]. Similarly, Yang and Wang corrupted One Billion Word Benchmark corpus by deleting a token, adding a token and replacing a token with the equal probability [58].

#### 5.1.2 Probabilistic Methods

Although deterministic noise injection methods are effective to some extents, however, the generated errors such as random word order shuffling and word replacing are less realistic than errors observed in GEC datasets. Many approaches inject noise to original data with consideration of more linguistic features, thus the generated errors resemble more to real errors made by English learners. We call these approaches probabilistic methods.

There are some typical works in prior stage. GenERRate [113] is an error generation tool with more consideration of POS, morphology and context. Rozovskaya and Roth proposed three methods to generate artificial article errors with more consideration of frequency information and error distribution statistics of ESL corpora: (1) injecting articles so that their distribution of generated data resemble distribution of articles in ESL corpora before annotator's correction; (2) injecting articles so that their distribution of generated data resemble distribution of articles in ESL corpora after annotator's correction; and (3) injecting articles according to specific condition probability. $P(token_{src}|token_{trg})$ is estimated from annotated ESL corpora, which means the probability that article tokensrc should be corrected into article $token_{trg}$. During error generation, article $token_{trg}$ in error-free context is replaced with $token_{src}$ with probability $P(token_{src}|token_{trg})$ [114]. Following this, inflation method [115] is proposed in order to solve the problem of error sparsity and low error rate in artificial data. This noise injection approach has been widely applied by following GEC models and systems, especially classification based [80], [81]. Yuan and Felice extracted two types of correction pattern from NUCLE corpus: context tokens and POS tags, and injected the patterns to EVP corpus with equal probability to generate artificial training data [24].

More recently, Felice and Yuan extended previous work by injecting errors with more linguistic information, and their approach generated 5 types of errors. Specifically, they refined the conditional probability $P(token_{src}|token_{trg})$ in following several aspects: (1) estimating the probability of each error type and use it to change relevant instances in error-free context; (2) estimating the conditional probability of words in specific classes for different morphological context; (3) estimating the conditional probability of source words when POS of target words are assigned; (4) estimating the conditional probability of source words when semantic classes of target words are assigned; (5) estimating the conditional probability of source words when the particular sense of target words are assigned. Experimental results validated the effectiveness of their consideration of various linguistic characteristics [116]. Based on this, the pattern extraction method [117] is an improvement and can be applied to generate errors for all types. In this method, correction patterns (*ungrammatical phrase, grammatical phrase*) are extracted from parallel sentences in training corpus, and errors are injected by looking for matches between error-free context and *grammatical phrase*. Xu et al. generated 5 types of errors, including concatenation, misspelling, substitution, deletion and transposition by considering the relation between sentence length and error numbers, and assigned a possibility distribution to generate each error type [100]. They designed an algorithm to create *word tree* that was used to create more likely substitution errors.

Other noise injection methods were also proposed and applied in recent GEC researches. Aspell spellchecker was applied to create confusion set to generate word replacement errors more accurately than randomly selecting a substitute from word list [57]. Choe et al. generated artificial errors basing on inspection of frequent edits and error types in annotated GEC corpora [55]. Besides, Kantor et al. generated synthetic data via applying the corrections observed in the training corpus W&I backward to native error-free data [105].

## 5.2 Back Translation

Back translation was initially proposed to augment training data for neural machine translation [118] and then adapted to GEC. In this kind of AEG methods, a reverse machine translation model is trained, translating the error-free grammatical data into ungrammatical data.

Rei et al. firstly trained a phrase-based SMT model on the public FCE dataset to generate errors. Trained model are further applied on EVP dataset to extract generated parallel data. They also made attempt on other monolingual (error-free) data such as Wikipedia dataset but failed to have equal quality data, which demonstrated by their development experiments where keeping the writing style and vocabulary close to the target domain gives better results compared to simply including more data [117].

Kasewa et al. firstly applied an attention-based neural encoder-decoder model on error generation [119]. Three different error generation methods were discussed and compared. (1) *Argmax* (AM) selects the most likely word at each decoding time step according to each candidate token's generation probability $p_i$. (2) *Temperature Sampling* (TS) involves a temperature parameter $\tau$ to alter the distribution:

$$\hat{p}_i = \frac{p_i^{1/\tau}}{\sum_j p_j^{1/\tau}}, \qquad (47)$$

which controls the diversity of generated tokens. (3) *Beam Search* (BS) maintains $n$ best hypothesis each time step according to their scores. Experimental results show that *Beam Search* error generation method improves the performance of error detection model trained on the generated parallel data most significantly.

However, the traditional beam search in decoder would result far fewer errors in generated ungrammatical sentences than original noisy text [103]. As a solution, the beam search noising schemes can be extended through (1) penalizing each hypothesis $\hat{y}$ in the beam by adding $k\beta_{rank}$ to their scores $s(\hat{y})$, where $k$ is their rank in descending log-probabilistic $p(\hat{y})$ and $\beta_{rank}$ is a hyperparameter (2) penalizing only the top hypothesis htop of the beam by adding $\beta_{top}$ to $s(h_{top})$ and (3) penalizing each hypothesis in the beam by adding $r\beta_{random}$ to their scores $s(\hat{y})$, where $r$ is a random variable between 0 and 1. As a result, more diversity and noise were involved in synthesised sentences.

*Quality control* [59] was applied to guarantee the generated sentences are less grammatical. To be more specific, the generated ungrammatical sentence $\tilde{y}$ can be added to training set only when $\tilde{y}$ and the error-free original sentence $y$ satisfy the following inequation:

$$\frac{f(y)}{f(\tilde{y})} \leq \sigma, \qquad (48)$$

where

$$f(y) = \frac{\sum_{t=1}^{T_y} \log p(y_t | y_{<t})}{T_y}. \qquad (49)$$

The variable $\sigma$ is learned on the development set with size $|N|$:

$$\sigma = \frac{\sum_{x,y}^{N} \frac{f(y)}{f(x)}}{N}. \qquad (50)$$

Htut and Tetreault compared several neural translation models that could be used on artificial error generation: multi-layer CNN, Transformer, PRPN [120], and ONLSTM [121]. Experiment results showed that ungrammatical sentences generated by multi-layer CNN and Transformer were more beneficial to training GEC systems. However, adding too much artificial data would impact the performance [122]. So, it is important to oversample the authentic training data to maintain the balance [91]. Besides the mentioned works, many other GEC systems also incorporate monolingual data via back translation [67], [99].

### 5.3 Others

Round-trip translation [107] is a parallel data generation method where the error-free original sentence is first translated into a bridge language, such as French, German, Japanese, etc. The translated sentences are then translated back into English as ungrammatical sentences. This is because noise is involved according to both the weaknesses of the translation models and the various inherent ambiguities of translation.

Cheat sheet [123] is a dictionary of error-focused phrases, which consists of two parts: (1) errors and their corresponding context that directly extracted from existing training data, and (2) mapping phrases with their translation probability over 95% from the SMT phrase table, which has a more random distribution than the prior over error location in a phrase.

### VI. Experiment

We have introduced public datasets, basic approaches, performance boosting techniques and data augmentation methods in GEC separately. It is time to summarize the experiment results of the introduced elements when they are applied in GEC systems, and present the analysis based on the experiment results. In this section, we first introduce the evaluation metrics in GEC. Then, we compare the properties and empirical performance of different GEC approaches. After that, we summarize the improved performance brought by the performance boosting techniques. At last, we recapitulate a wide range of GEC systems whose approaches, performance boosting techniques and data augmentation methods have been introduced in this survey for a more clear pattern of current GEC works.

TABLE 5
Different evaluation metrics

| Metric | Definition | Multiple References | System | | Sentence |
|--------|-----------|---------------------|--------|--------|----------|
| | | | Pearson $r$ | Spearman $\rho$ | Kendall $r$ |
| $M^2$ | Eq 54- 56 | max | 0.623 | 0.687 | 0.617 |
| $I$ | Eq 58- 59 | max | -0.25 | -0.385 | 0.564 |
| GLEU | Eq 60- 62 | average | 0.691 | 0.407 | 0.567 |
| ERRANT | Eq 54-56 | max | 0.64 | 0.626 | 0.623 |

### 6.1 Evaluation

Appropriate evaluation of GEC has long been a hot issue [106], [16], [124], [125], due to the subjectivity, complexity and subtlety of GEC [126]. Generally, evaluation methods in GEC can be divided into two branches: reference-based and reference-less. The difference is whether reference is required during evaluation. Since reference-based methods are more commonly applied, we concentrate mainly on this type of metrics.

#### 6.1.1 Reference-based Metrics

Reference-based evaluation metrics are computed by comparing the hypotheses with references or golden standards. Since traditional precision, recall and F-score can be misleading [127], [15], some evaluation metrics have been proposed, including $M^2$ [128], $I$ [129], GLEU [126] and ERRANT [11]. When multiple references are available, $M^2$, $I$ and ERRANT choose the one that maximize the metric score, while GLEU randomly selects one reference and calculates the average of 500 scores as the final score.

Some works compared these metrics in relation to human judgements in both system-level and sentence-level [7], [130]. System-level correlation coefficient is calculated by comparing the ranking of the systems by human evaluation and the ranking generated by the metric scores, while sentence-level correlation efficient is computed by examining the number of pairwise comparisons that metrics agree or disagree with humans. The first is system-level Pearson ($r$):

$$r = \frac{\sum_{i=1}^{q}(a_i - \bar{a})(b_i - \bar{b})}{\sqrt{\sum_{i=1}^{q}(a_i - \bar{a})^2}\sqrt{\sum_{i=1}^{q}(b_i - \bar{b})^2}}, \qquad (51)$$

where $a$ and $b$ are system scores given by metric and human respectively. The variable $q$ is the number of GEC systems. The second is system-level Spearman ($\rho$):

$$\rho = 1 - \frac{6\sum_{i=1}^{q}(d_i)^2}{q(q^2 - 1)}, \qquad (52)$$

where $d$ is the difference between metric rank and human rank. The third is sentence-level Kendall's Tau ($\tau$):

$$\tau = \frac{|Concordant| - |Discordant|}{Total \ \# \ Pairwise \ Comparisons}, \quad (53)$$

where $|$Concordant$|$ and $|$Discordant$|$ are the numbers of pairwise comparisons that metric agrees and disagrees with human. Comparison of different metrics is listed in Table 5.

$\boldsymbol{M^2}$. MaxMatch ($M^2$) is the most commonly used evaluation metric in GEC today. $M^2$ relies on error-coded annotations of golden standard. Phrase-level edits of system hypothesis are extracted by $M^2$ with the maximum overlap with the golden standard. The edits are evaluated with respect to gold edits in $F_\beta$ measure. Suppose the extracted set of gold edits for sentence $i$ is $g_i$, while the set of system hypothesis edits for sentence $i$ is $e_i$. Recall $R$, precision $P$, and $F_\beta$ are defined as follows:

$$R = \frac{\sum_{i=1}^{n} |g_i \cap e_i|}{\sum_{i=1}^{n} |g_i|}, \quad (54)$$

$$P = \frac{\sum_{i=1}^{n} |g_i \cap e_i|}{\sum_{i=1}^{n} |e_i|}, \quad (55)$$

$$F_\beta = \frac{(1 + \beta^2) \times R \times P}{R + \beta^2 \times P}, \quad (56)$$

where

$$g_i \cap e_i = \{e \in e_i | \exists g \in g_i, match(g, e)\}, \quad (57)$$

where $match(g, e)$ means that $e$ and $g$ have the same offset and correction. $F_{0.5}$ emphasizes precision twice as much as recall, and precision is more important than recall in GEC. This is because it is more desirable that edits of system hypothesis are actually grammar errors than some of the edits are erroneous correction.

$\boldsymbol{I}$. $M^2$ has a series of disadvantages. First, it does not discriminate do-nothing systems and erroneous correction proposing systems, since their $F_{0.5}$ will all be 0. Second, phrase-level edits may not always reflect effective improvements, thus misleading evaluation outcomes. Besides, error detection scores are not computed. Aiming at addressing the disadvantages, *Improvement* measure, $I$, is proposed. Based on the alignment among the source sentence, hypothesis and golden standard, $I$ is computed using weighted accuracy where TPs and FPs are weighted higher than TNs and FNs. Specifically, given the aligned tokens $\omega^{src}$, $\omega^{hyp}$ $\omega^{std}$ in source, hypothesis and golden standard, TP, FP, TN, FN are defined as follows:

- TP: $\omega^{src} \neq \omega^{std}$ while $\omega^{hyp} = \omega^{std}$;
- FP: $\omega^{src} \neq \omega^{hyp}$ while $\omega^{hyp} \neq \omega^{std}$;
- TN: $\omega^{src} \neq \omega^{hyp} = \omega^{std}$;
- FN: $\omega^{src} \neq \omega^{std}$ while $\omega^{hyp} \neq \omega^{std}$.

The weighted accuracy is calculated as follows:

$$WAcc = \frac{\lambda \cdot TP + TN}{\lambda \cdot TP + TN + \lambda \cdot (FP - \frac{FPN}{2}) + (FN - \frac{FPN}{2})}, \quad (58)$$

where FPN are cases that can be classified as both FP and FN. $\lambda$ is set to 2 to give TPs and FPs higher weights. The metric $I$ is calculated according to:

$$I = \begin{cases} \lfloor WAcc \rfloor, & if WAcc = WAcc_{src} \\ \frac{WAcc - WAcc_{src}}{1 - WAcc_{src}}, & if WAcc > WAcc_{src} \\ \frac{WAcc}{WAcc_{src}} - 1, & otherwise. \end{cases} \quad (59)$$

where $WAcc_{src}$ is calculated by treating source sentences as hypothesises as well.

**GLEU.** Both $M^2$ and I require explicit error annotations. Inspired by machine translation, GEC can be treated as sequence-to-sequence rewriting and propose Generalized Language Evaluation Understanding metric (GLEU). GLEU calculates weighted precision of n-grams of hypothesis over references, which rewards correctly changed n-grams while penalizing n-grams that appear in the source sentence but not in the references. In the experiment, GLEU has the strongest correlation degree with human evaluation. To calculate GLEU score, n-gram precision is firstly calculated as follow:

$$p_k = \frac{\sum_{\hat{y}^i \in \hat{Y}} \left[ \sum_{n \in N} \#_{\hat{y}^i, y^i} n - \sum_{n \in N} B(\#_{\hat{y}^i, x^i} n - \#_{\hat{y}^i, y^i} n) \right]}{\sum_{\hat{y}^i \in H} \sum_{n \in N} \#_{\hat{y}^i} n}, \quad (60)$$

where $\hat{Y}$ is a set of corrected hypotheses, $B(x)$ is the bigger number between 0 and $x$, $\#_a n$ is the number of n-gram sequences in $a$, and $\#_{a,b} n$ is the minimal number of n-gram sequences in $a$ and $b$. $N = \{1,2,3,4\}$. The brevity penalty is computed according to

$$BP = \begin{cases} 1, & if \ l_h > l_r \\ \exp(1 - l_r/l_h), & if \ l_h \leq l_r \end{cases}, \quad (61)$$

where $l_r$ is token number of the references and $l_h$ is the token number of all hypotheses. Then, GLEU is finally calculated as

$$GLEU(S, \hat{Y}, R) = BP \cdot \exp(\frac{1}{N} \sum_{k=1}^{N} \log p_k), \quad (62)$$

where $S$ and $R$ are sets of source sentences and references respectively.

**ERRANT.** ERRANT is an improved version of $M^2$ scorer. It first extracts the edits, then classifies the errors into 25 categories. ERRANT $F_\beta$ is calculated as same as $M^2 F_\beta$. While ERRANT and $M^2$ both evaluate span-based correction, ERRANT also reports the performance on spanbased detection and token-based detection. Besides, it is the first metric that is capable of evaluating the performance on different error types, which is more beneficial for development of GEC systems.

### 6.1.2 Reference-less Metrics

Reference-less evaluation is not researched until recently. The motivation is that reference-based metrics always penalize edits that are not included in reference, resulting into unfair underestimating of GEC systems. Several reference-less evaluation metrics are researched. The first trial is the research about LFM score [131] and error count score given by e-rater (ER) and Language Tool (LT) [132]. Then, adding fluency and meaning preservation into consideration demonstrates that reference-less metrics can replace reference-based metrics in evaluating GEC systems [133]. Besides, a statistical model is trained to scale the grammaticality of a sentence using features including misspelled counts, n-gram language model scores, parser outputs, and features extracted from precision grammar parser [131]. Finally, based on UCCA scheme [134], $US_{IM}$ is a reference-less measure of semantic faithfulness which compares the differences between the UCCA structures of outputs and source sentences according to alignable but different tokens. Comparing to RBMs requiring many references, $US_{IM}$ has a lower cost in GEC systems. It also shows sensitivity for changes in meaning [135].

### 6.2 Analysis of Approaches
This section presents analysis of previous discussed approaches, including approaches based on SMT, NMT, LM and classification. The analysis mainly focuses on two aspects: the development of MT based approaches and the comparison among different approaches.

#### 6.2.1 Development of Approaches

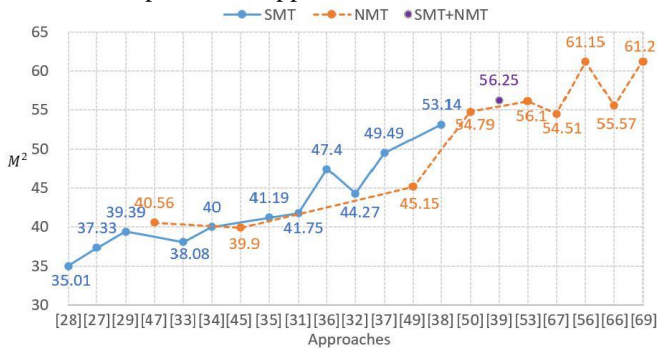

Fig. 4. Development of SMT based approaches and NMT based approaches.

We summarize the experimental results of SMT based systems and NMT based systems that we have introduced in Section 3 in Figure 4. The systems are evaluated on CoNLL-2014 test set with $M^2$ metric. The experiment results of the selected systems majorly reflect the performance of the approaches themselves, although some may be combined with other performance boosting techniques. The systems are arranged in chronological order. It is obvious that incremental performance is achieved in both SMT based approaches and NMT based approaches.

The potential of SMT based GEC systems were first exposed through their top tier performance in CoNLL-2014 shared tasks with 35.01 and 37.33 of $F_{0.5}$ score [28], [27]. Following work [37] developed models tailored to GEC task by incorporating GEC-specific features into standard SMT based approaches and tuning towards $M^2$.

Their system proposed a strong SMT baseline at a 49.49 $F_{0.5}$ score even outperformed previous hybrid approaches [29], [36] that made attempts to take advantage of both SMT and classifiers. The integration of neural network features such as NNJM and NNGLM in SMT based system bring benefit [31], [32], [38]. Apart from incremental performance produced by NNJM, the 53.14 $F_{0.5}$ score was achieved due to an additional character-level SMT model [38]. It is also confirmed that systems augmented with reranking technique enjoy extra bonus in final performance [33], [34], [35], which require no modification in model structures.

For NMT model, combining the word-level attention and character-level attention proves effective, achieving 45.15 $F_{0.5}$ [49]. The multi-layer CNN model brings significant improvement, but the final performance (54.79 $F_{0.5}$) is not only achieved by the translation model itself, but also enhanced by performance boosting techniques by about 8 points [50]. Replacing traditional RNN with Transformer and other performance boosting techniques also behaves better [53]. Another significant improvement is attained by adding copy mechanism to pre-trained Transformer [56]. We will discuss more details about the performance boosting techniques in the next section.

Apart from the analysis above concerning SMT and NMT separately, we could also draw a brief conclusion from those broken lines in Figure 4 with regard to several turning points and preference switches among different approaches. Before the CoNLL-2014 tasks, classification based and rule based approaches are widely used in GEC, which are not involved in the Figure 4 in advance of all works we described in the lines graph. Although SMT was proposed earlyin2006 [22] with considerable performance in a certain error type, further researches stagnated due to the lack of parallel data until those shared tasks together with the propose of NUCLE and utilizing web-crawled language learner data in GEC [23],[24]. Systems exploiting SMT based approaches [28], [27] obtained 35.01 and 37.33 respectively in CoNLL-2014 shared task as shown in the graph, seizing the first and third place among all systems. Later work [29] that combined both SMT and classification in 2014 obtained the state-of-the-art performance with 39.39 $F_{0.5}$ score, which demonstrated that SMT based methods had achieved the equal or even more significant position compared to classifiers in the field of GEC. More researches employing SMT in 2016 developed models that better tailored with GEC task and a strong baseline were created entirely with SMT [37], which reached a new state-of-the-art result 49.49. This work also symbolized the peak of SMT based systems in GEC then. Early in the same year, several pilot works introduced NMT based methods, utilizing attention

TABLE 6

Comparison of different GEC approaches. We use "+" and "-" to indicate the positive and negative characteristics of the approaches with respect to each property.

| Property | SMT | NMT | Classification | LM |
|---|---|---|---|---|
| Error coverage | + all errors occurring in the training data | + all errors occurring in the training data | - only errors covered by the classifiers | - only errors covered by the confusion set |
| Error complexity | + automatically through parallel data; limited inside a phrase | + automatically through parallel data; capture longer dependency | - need to develop via specific approaches | - need to develop more complex confusion sets |
| Generalize ability | - only confusions observed in training can be corrected | + generalize through continuous representation and approximation | + well generalizable via confusion sets and features | - only errors in confusion sets can be corrected |
| Supervision/Annotation | - required | - required | + not required | + not required |
| System flexibility | - inflexible; sparse feature used for specific phenomenon | - inflexible; multi-task learning facility specific phenomenon | + flexible; phenomenon-specific knowledge sources | + flexible; injecting knowledge into confusion sets |

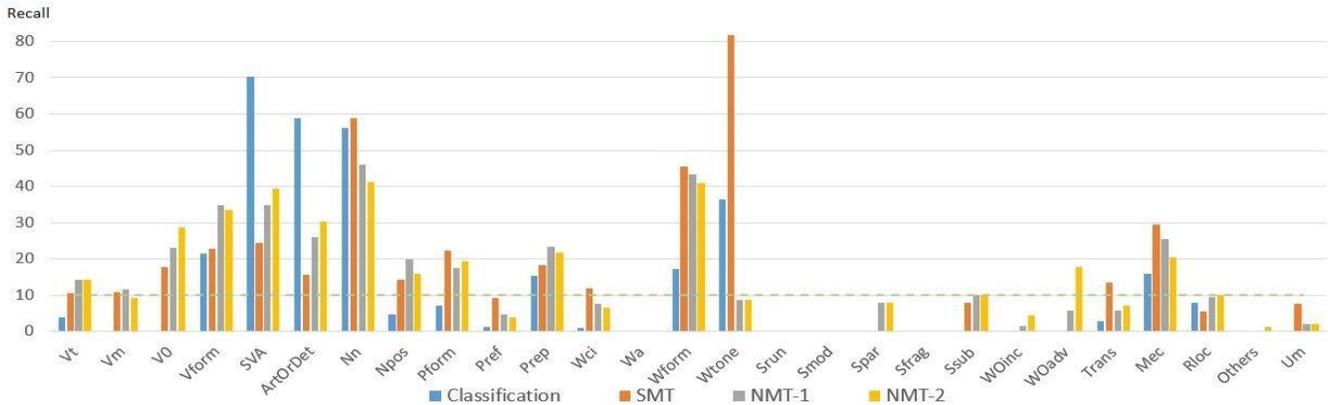

Fig. 5. Recall for each error type, indicating how well different approaches perform against a particular error type. The green imaginary line is at the height where recall is 10. Both the classification and SMT systems we select from the CoNLL-2014 shared task are almost bare-bone systems without any other performance boosting techniques. The NMT-1 and NMT-2 represent systems using a CNN-based and an RNN-based sequence-to-sequence model respectively. Results of LM-based methods are not covered due to the lack of relative empirical works.

augmented RNN, into GEC task [47], [45]. Although those NMT based methods were never on trial in GEC before, they still showed promising results in GEC task with 40.56 and 39.9 $F_{0.5}$ score. Under the influence of the empirical performance of neural methods, systems based on SMT further modified their models with additional neural features to remedy the defect of SMT methods that are unable to capture context information or make continuous representations. Indeed, a range of study [31], [32], [38] demonstrated that neural network features such as NNJM did create extra improvement to standard SMT model in GEC.

More recently, NMT based model that employed convolutional sequence-to-sequence model holding an $F_{0.5}$ score of 54.79 outperformed predominated SMT approaches [50], in which dependency that correction process required could be more readily to capture than RNN by constructing shorter link path among tokens. The developed techniques and optimized network structures better released the potential of neural methods on incorporating context and prompted the NMT based approaches to a new state-of-the-art. Following hybrid system [39] that combined NMT and SMT also proved the state and efficiency of NMT based approaches in GEC. Then transformer, competent of constructed more flexible and efficiency links, were introduced into GEC and produced better performance [53], after which most of GEC systems preferred transformer as a baseline model. Besides, work using a parallel iterative model, inspired by the using of parallel models in machine translation, also obtained a comparative performance with those encoder-decoder models [69]. Benefiting from its advantage in time cost and GEC tailored structures, similar model framework deserves further development.

### 6.2.2 Comparison Among Approaches

We now present the comparison of different approaches from several key properties that distinguish among GEC approaches and that we identified to better characterize those models' frameworks as well as output performance. We majorly consider properties as previous empirical work [36] does, but extend their conclusion to all types of models in GEC:

- **Error coverage** denotes the ability of a system to identify and correct a variety of error types.
- **Error complexity** indicates the capacity of a system to address complex mistakes such as those where multiple errors interact.
- **Generalize ability** refers to the ability of a system to identify mistakes in new unseen contexts and propose corrections beyond those observed in training data.
- **The role of supervision** or having annotated learner data for training.
- **System flexibility** is a property of the system that allows it to adapt resources specially to correct various phenomena.

The first three items reflect the property of system output while the other two items characterize more aspects of system frameworks. All conclusions of analysing those four types of approaches have been demonstrated in Table 6, while more detailed information will be covered below.

We first employ error coverage to understand how systems differ in capturing patterns of various error types and its corresponding corrections. Similar to previous work [36], we regard recall of each error categories as a reflection of systems' competent to detect different types of errors. More specifically, if a system has performance with non-negligible recall (higher than 10%) on a certain error type, we consider such error category is covered by this system. To draw our conclusion, a range of systems are compared together concerning their recall numbers with several experimental studies scrutinized. The classification based methods and LM based methods are limited in covering more error categories due to the classifiers or confusion sets that often tailor with specific error type. On the contrary, almost all types of errors are well-covered by MT based methods (both SMT and NMT), in which the error patterns are automatically captured by training on the error-complete parallel corpus. It is also worth noticing that high recall does not mean high precision, especially in SMT. An SMT phrase-table ensures the coverage by building all possible correction mapping which also gives high probability to mis-correction demonstrated by precision numbers.

To handle with condition that an error type involves complex correction as well as interaction among multiple error types, different methods employ distinct strategies. In SMT based approaches, all the information of errors and its corresponding correction is implicated inside the phrase-table obtained from parallel datasets, which also limits their correction pattern can only be done inside a phrase and context as well as other global information beyond a single phrase is often ignored. In NMT based approaches, context information is often considered and links among multiple tokens are capable to build. Although long dependency problem still remains when RNN or LSTM is used, model with more flexible links like convolutional sequence-to-sequence model and Transformer have provided better solutions. Those NMT models outperform previous SMT based methods partially because of the benefit of shorter link path among tokens. However, LM and classification based methods can hardly achieve an equal performance in dealing such complexity with those generative models due to the limitation of both classifiers and confusion sets.

TABLE 7
Improvement brought by the techniques.

| Performance Boosting Technique | Source | Test Set | Metric | Before | After |
|---|---|---|---|---|---|
| Pre-training | [56] | CoNLL-2014 | $M^2$ | 54.67 | 58.80 |
| | [100] | W&I+L dev | ERRANT | 47.01 | 55.37 |
| | [100] | W&I+L test | ERRANT | 53.24 | 67.21 |
| | [53] | CoNLL-2014 | $M^2$ | 51.00 | 54.10 |
| Reranking | [66] | CoNLL-2014 | $M^2$ | 54.25 | 55.57 |
| | [59] | W&I+L dev | ERRANT | 49.68 | 50.39 |
| | [57] | W&I+L dev | ERRANT | 52.30 | 53.00 |
| | [61] | W&I+L dev | ERRANT | 32.60 | 35.35 |
| | [102] | CoNLL-2014 | $M^2$ | 55.72 | 55.77 |
| | [39] | CoNLL-2014 | $M^2$ | 53.51 | 54.95 |
| | [50] | CoNLL-2014 | $M^2$ | 49.33 | 54.13 |
| | [104] | CoNLL-2014 | $M^2$ | 49.34 | 51.08 |
| Ensemble | [56] | CoNLL-2014 | $M^2$ | 59.75 | 61.15 |
| | [54] | CoNLL-2014 | $M^2$ | 54.90 | 58.30 |
| | [54] | JFLEG | GLEU | 59.30 | 62.40 |
| | [100] | CoNLL-2014 | $M^2$ | 60.90 | 63.20 |
| | [100] | JFLEG | GLEU | 60.80 | 62.60 |
| | [55] | CoNLL-2014 | $M^2$ | 63.05 | 69.06 |
| | [91] | CoNLL-2014 | $M^2$ | 55.72 | 58.90 |
| | [90] | CoNLL-2014 | $M^2$ | 56.33 | 58.40 |
| | [90] | JFLEG | GLEU | 55.39 | 55.60 |
| | [66] | CoNLL-2014 | $M^2$ | 53.06 | 54.25 |
| | [50] | CoNLL-2014 | $M^2$ | 46.38 | 49.33 |
| Auxiliary Task Learning | [56] | CoNLL-2014 | $M^2$ | 58.80 | 59.76 |
| | [108] | W&I+L test | ERRANT | 59.88 | 60.97 |
| | [99] | W&I+L test | ERRANT | 69.80 | 70.20 |
| Iterative Correction | [54] | CoNLL-2014 | $M^2$ | 55.20 | 58.30 |
| | [54] | JFLEG | GLEU | 57.00 | 62.40 |
| | [84] | W&I+L dev | ERRANT | 53.05 | 53.72 |
| | [60] | W&I+L dev | ERRANT | 43.29 | 44.27 |
| Edit-weighted Objective | [53] | CoNLL-2014 | $M^2$ | 48.00 | 51.00 |

All the correction produced by MT systems (both NMT and SMT based methods) are derived from patterns appeared in parallel corpus. The learned rules or grammatical information they obtained is inducted from the surface form of the error-correction pairs, which also leads to the difficulty of generalization. In SMT methods, the mined phrase-table almost places such a restriction on the correction that only correction they have met could be produced. The LM based approach also confronted with the similar circumstance as the phrase table did in SMT based methods due to the mapping constructed by confusion sets. The neural methods partially handled this problem by employing a continuous representation space

TABLE 8

Recapitulation of integrated GEC systems. "A" and "E" refer to auxiliary task learning and edit-weighted object respectively. We make the following explanation with regard to systems notated with asterisk. Round-trip translation [107] is described in Section 5.3. Approaches of systems notated with three asterisks are language model based or classification based. Training these systems requires large monolingual instead parallel training data, so their grids in the column of "Corpus" are blank.

| Source | Year | Approach | | | | | Performance Boosting Technique | | | | | Data Augmentation | | Corpus | | | | Evaluation | | |
|---|---|---|---|---|---|---|---|---|---|---|---|---|---|---|---|---|---|---|---|---|
| | | SMT | NMT | Classification | LM | Hybrid | Pre-training | Reranking | Ensemble | Iterative Correction | Others | Noising | Back Translation | NUCLE | FCE | Lang-8 | Wiki+ LOCNESS | CoNLL-2014 M² | BEA-2019 ERRANT | JFLEG GLEU |
| [99] | 2019 | | √ | | | | √ | √ | √ | | | √ | √ | | | | √ | 65 | 70.2 | 61.4 |
| [57] | 2019 | | √ | | | | √ | | √ | | | √ | √ | | | | √ | 64.16 | 69.47 | 61.16 |
| [100] | 2019 | | √ | | | | | | √ | | | | √ | | | | √ | 63.2 | 66.61 | 62.6 |
| [69]* | 2019 | | √ | | | | √ | | √ | √ | | | | √ | √ | √ | | 61.2 | - | 61 |
| [56] | 2019 | | √ | | | | √ | √ | √ | | A | | | √ | √ | √ | | 61.15 | - | 61.5 |
| [107]** | 2019 | √ | | | | | | | √ | √ | | | | | | | √ | 60.4 | - | 63.3 |
| [55] | 2019 | | √ | | | | | | √ | | | √ | | √ | √ | √ | | 60.33 | 69 | - |
| [91] | 2019 | | √ | | | | | | √ | | | | √ | | | | √ | 58.9 | 63.72 | - |
| [90] | 2019 | √ | √ | | | √ | | | √ | | | | | | | √ | √ | 58.4 | - | 55.6 |
| [66] | 2019 | | √ | | | | √ | √ | √ | | | | | √ | | √ | | 55.57 | - | - |
| [105] | 2019 | | √ | | | | √ | | √ | | | √ | | √ | √ | √ | | - | 73.18 | - |
| [84] | 2019 | | √ | | | | √ | | √ | √ | | | | √ | √ | √ | | - | 66.78 | - |
| [59] | 2019 | | √ | | | | √ | | | | A | | | √ | √ | √ | | - | 66.75 | - |
| [108] | 2019 | | √ | | | | | | | | A | | √ | √ | √ | √ | | - | 60.97 | - |
| [60] | 2019 | | √ | | | | | | | | E | | | √ | √ | √ | | - | 59.39 | - |
| [58] | 2019 | | √ | | | | | | | √ | E | √ | | √ | √ | √ | | - | 58.62 | - |
| [61] | 2019 | | √ | | | | √ | | | | | | | √ | √ | √ | | - | 53.45 | - |
| [92]*** | 2019 | | | √ | | | | | | | | | | | | | | - | 40.17 | - |
| [54] | 2018 | | √ | | | | √ | | √ | | | | | | | √ | | 58.3 | - | 62.4 |
| [102] | 2018 | | √ | | | | √ | √ | | | E | | | √ | | | | 56.52 | - | - |
| [39] | 2018 | √ | √ | | | √ | √ | | | | | | | √ | | √ | | 56.25 | - | 61.5 |
| [53] | 2018 | | √ | | | | √ | | | | E | | | √ | | | | 55.8 | - | 59.9 |
| [50] | 2018 | | √ | | | | √ | | √ | | | | | | | √ | | 54.79 | - | 57.47 |
| [67] | 2018 | | √ | | | | √ | | | | | | √ | | | √ | | 54.51 | - | 57.74 |
| [86]*** | 2018 | √ | | √ | | | | | √ | | | | | | | | | 50.16 | - | - |
| [89]*** | 2018 | | | √ | | | | | | | | | | | | | | 34.09 | - | 48.75 |
| [68] | 2018 | | √ | | | | | | | | | | | √ | √ | √ | | - | - | 53.98 |
| [38] | 2017 | √ | | | | | | | | | | | | √ | √ | √ | | 53.14 | - | 56.78 |
| [104] | 2017 | √ | | | | | | | | | | | | √ | | √ | | 51.68 | - | 43.26 |
| [49] | 2017 | | √ | | | | | | | | | | | √ | | √ | | 45.15 | - | - |
| [83]*** | 2017 | | | | √ | | | | | | | | | | | | | 41.6 | - | - |
| [37] | 2016 | √ | | | | | | | | | | | | | | | | 49.49 | - | - |
| [36] | 2016 | √ | | √ | | √ | | | | | | | √ | | | | | 47.4 | - | - |
| [32] | 2016 | √ | | | | | | | | | | | | | | √ | | 44.27 | - | - |
| [31] | 2016 | √ | | | | | | | | | | | | | | √ | | 41.75 | - | - |
| [35] | 2016 | √ | | | | | | | √ | | | | | | | | | 41.19 | - | - |
| [34] | 2016 | √ | | | | | | | √ | | | | | | | | | 40 | - | - |
| [33] | 2016 | √ | | | | | | | | | | | | | | √ | | 38.08 | - | - |
| [47] | 2014 | √ | | | | | | √ | | | | | | | | | | 40.56 | - | - |
| [45] | 2014 | √ | | | | | | | | | | | | | | √ | | 39.9 | - | - |
| [29] | 2014 | √ | | | | | | | | | | | | | | | | 39.39 | - | - |
| [27] | 2014 | √ | | √ | | √ | | √ | | | | √ | | | | √ | | 37.33 | - | - |
| [81] | 2014 | √ | | | | | | | | | | | | | | | | 36.79 | - | - |
| [28] | 2014 | √ | | | | | | | | | | | | | | √ | | 35.01 | - | - |

and their approximation property when learning a pattern, nevertheless limitation still remains when we require further generalization. Although the classification cannot well cover all type of errors, it still demonstrates its competent of generalization in dealing with certain error types. For category such as prepositions or articles, several more general rules are obtained.

Another property of model framework is flexibility that whether the model allows for a flexible architecture where a range of error-specific knowledge could be incorporated by tailoring with additional resources. The fact that classification methods exploiting various classifiers suited to individual error types has demonstrated the flexibility of those methods by adjusting or augmenting with specific classifier. Although the MT based methods appear to have a more fixed structure where a whole model is used to solve all error types, several additional techniques have been proposed to better incorporate extra resources based on the standard MT methods. For SMT, often a range of features are integrated in log linear model, where such architecture also allows incorporation of extra information as sparse features such as syntactic patterns and certain correction patterns. Whereas in NMT based methods, extra information is more difficult to be injected, although methods such as multitask learning could be used directly against several specific phenomenons.

## 6.3 Analysis of the Techniques

Table 7 summarizes the improvement of system performance brought by the performance boosting techniques that we have introduced in Section 4. The experiment results come from the ablation study in their original paper and we collect them together. Most systems are evaluated on CoNLL-2014 test set using $M^2 F_{0.5}$ evaluation metric.

While all the techniques indeed lift the performance in some degree, pre-training is the most effective, especially when considering the significant increase in ERRANT $F_{0.5}$ on W&I+LOCNESS development set, from 47.01 to 55.37 with an increase of 8.36, and test set, from 53.24 to 67.21 with an increase of 13.97 [100]. Ensemble and reranking both yield steady improvement, and ensemble is generally a little more beneficial than reranking. Auxiliary task learning brings the least improvement, possibly because the systems have already been combined with other techniques and enhanced, weakening the additional benefit brought by auxiliary task learning only. Although iterative correction seems promising on CoNLL-2014 test set, with an increase of 3.10 $M^2 F_{0.5}$, the beneficial effect is minor on W&I+LOCNESS development set (0.67 ERRANT $F_{0.5}$ [84] and 0.98 ERRANT $F_{0.5}$ [60]). Besides, the boosting effect of edit-weighted object (3.00 $M^2 F_{0.5}$) on the only system [53] may not be generalized to other systems.

## 6.4 Analysis of GEC Systems

Table 8 recapitulates typical GEC systems after CoNLL-2014 plus top 3 ranking systems in CoNLL-2014, where each column of the table corresponds to a subsection of the survey. Systems include works that present initial trial on new approaches, researches that explore new beneficial techniques and data augmentation methods, and submissions to CoNLL-2014 and BEA-2019 shared tasks comprising different components based on existing works. Note that for CoNLL-2014 shared task, we include only 3 top-ranking systems, since the performance of most systems in CoNLL2014 are relatively limited compared to subsequent works, weakening their value for reference. For BEA-2019 task, a system might participate into multiple track, so we include the evaluation results on restricted track, where the available annotated training data is the same as other systems (i.e. the FCE, Lang-8, NUCLE and W&I+LOCNESS) and the unlimited monolingual data is the same as other systems, for more controlled analysis. For a whole report on participants in BEA-2019, we refer to the task overview paper [7]. We summarize each system in aspects of what basic approaches they are based on, what techniques and data augmentation methods they used, and what datasets are used for training. Systems are evaluated on at least one of the three test sets with their corresponding evaluation metric: CoNLL-2014 test set with $M^2 F_{0.5}$, BEA-2019 test set with ERRANT $F_{0.5}$, and JFLEG with GLEU.

We include the best score and their corresponding configuration of approaches, techniques and data augmentation methods. Systems before CoNLL-2014 shared task are not included, since they are not evaluated on any one of the three test sets. Systems are ordered in primary consideration of their published year. For systems in the same year, we rank them according to their performance on CoNLL-2014 test set first, then BEA-2019 test set, and at last JFLEG. This is because CoNLL-2014 test set is the most widely used benchmark in the GEC research community, whereas BEA-2019 test set and JFLEG are not adopted as frequently. All systems evaluated on BEA-2019 test set using ERRANT metric, except 2 systems [99], [105], are submissions to the BEA-2019 shared task, among which some also report $M^2$ on CoNLL-2014 test set while others do not. Besides, no system is evaluated on JFLEG before it was released.

Great progress has been achieved in the past 5 years after CoNLL-2014 shard task. Although submissions to CoNLL-2014 shared task we included in Table 8 ranked top 3 [27], [28], [81], they are largely superseded by more powerful systems. Incremental improvement on performance is observed when it comes closer to 2019. Several systems participated in BEA-2019 are also evaluated on CoNLL-2014 test set using $M^2$ evaluation metric [99], [57], [100], [55], [91], and their scores have significantly surpassed the scores of top-ranked participants in CoNLL-2014 shared task.

Neural machine translation based approaches have been increasingly popular in recent years thanks to the popularity of deep learning. It is obvious that NMT based approach has become dominant since 2017. All the submissions to BEA-2019 shared task are NMT based, except for [92], participating in low-resource track only and adopting language model based approach. However, although NMT based GEC was researched as early as 2014 [45], [47], it does not outperform state-of-the-art SMT based GEC until [50]. With the propose of Transformer, many NMT based systems start to gain higher scores. Nearly two thirds of systems in BEA-2019 shared task are based on Transformer, which also lead to similar scores for some NMT based systems [100], [84], [59].

The proposal and application of performance boosting techniques also contribute to the better performance of GEC systems. Pre-training has become the requisite for more powerful NMT based GEC systems no matter what other techniques are adopted, for the sake of its unmatched ability to incorporate artificial training data or large monolingual data, for example, systems with the top 5 scores on CoNLL2014 test set [99], [57], [100], [69], [56], and systems ranking top 4 on BEA-2019 test set [105], [99], [57], [55]. Ensemble is the most widely adopted technique and always yields better performance than single model. More than half of the systems in 2019 used the combination of NMT based GEC approach, pre-training and ensemble of multiple basic model with identical architecture. It is worthy to mention that system with the highest ERRANT score (73.18) [105] and system with the fourth highest score (66.78) [84] on BEA2019 test set are based on sophisticated ensemble strategies combining the output of different GEC models, highlighting the benefit brought by combining different GEC systems. We have discussed them in Section 4.3. While pre-training and ensemble are restricted in NMT based systems, reranking is more universal since it is more model-independent. Reranking has been researched as early to 2014 and combined with SMT based systems [27], [33], [34], [35]. Besides, reranking is also important for better final performance, as it is adopted by the top 2-5 systems evaluated on BEA-2019 test set. Iterative correction is not as widely applied and effective, but can be also used for systems based on more approaches [69], [86]. Auxiliary task learning and edit-weighted training object are less extensively used and restricted with NMT based systems.

An important discrepancy between the two data augmentation methods is that data generated via noising is always used for pre-training [99], [57], [100], [56], [55], [105], while data generated by back translation is viewed as equivalent as authentic training data. This is because data generated using noising may not precisely capture the distribution of error pattern in realistic data, although many explorations are conducted to improve the quality of the noisy data (section 5.1). Besides, it can be concluded that as the noising strategy becomes more sophisticated, the better performance is gained, for example, the system pre-trained on the augmented data generated by sophisticated process achieved 63.2 $M^2F_{0.5}$ score on CoNLL-2014 test set [100], about 5 points higher than the system pre-trained on the data generated by deterministic approach, which achieved 58.3 $M^2F_{0.5}$ on CoNLL-2014 test set [54].

MT based systems rely heavily on parallel training data. Almost all the participants in BEA-2019 used all the training corpora available, and NUCLE and Lang-8 are used by nearly every MT based systems. FCE is not used as extensively, possibly due to the lack of multiple references. Although the performances of classification based and LM based systems are relatively limited compared to MT based systems, the nature of their low reliance on parallel training data adds many advantages to them for the sake of better application on low-resource scenario, for example, grammar error correction for other languages instead of English. In aspect of evaluation, no system obtains the highest score on all test sets according to their corresponding metric. This is suggestive that evaluation results may vary with different test sets, about which we discuss more in Section 7.

## VII. PROSPECTIVE DIRECTION

In this section, we discuss several prospective directions for future work.

- Adaptation to L1. Most of the existing works are L1 agnostic, treating text written by writers with various first language as equivalent. However, for better application of GEC, prospective GEC systems should provide more individualized feedback to English learners, taking consideration of their L1s. Although some works have already focused on several specific first language [32], [136], there is still much room to be explored.

- Low-resource scenario GEC. Datasets in machine translation have tens of millions sentence pairs, but the amount of parallel data in GEC is not comparable even in the largest corpus Lang-8. What's worse, when applied to other minority languages where a large amount of parallel data is not available, the performance of many powerful MT based GEC systems will be seriously impaired. So, training better GEC systems in low-resource scenario is waiting to be explored. The possible solution may be better pretraining and data augmentation strategies to incorporate large error-free text, and more exploration on LM based approaches without the reliance on supervision.

- Combination of disparate systems. It is demonstrated that different GEC systems are better at different sentence proficiency [7]. Besides, as indicated before, the system achieving the 1st and 4th best performance on BEA-2019 test set are based on ensemble of multiple individual GEC systems. These evidences suggest that it is promising to explore combination strategies to better incorporate the strength of disparate GEC systems, which may be specialized at different error types, topics, sentence proficiency extents and L1s.

- Datasets. The most widely used dataset in GEC is NUCLE, which has simplex sentence proficiency, topic and L1, as indicated in Table 1. The lack of variety means that the performance of GEC systems in other conditions remains unknown [137]. Besides, most datasets contain only limited number of references for source sentences. More references bring increase in inter-annotator agreement [15]. However, adding references requires extra labor. It is waiting to be researched that how many references should be added to datasets with consideration of cost.

- Better evaluation. Systems have always been evaluated on a single test set. However, different test sets lead to inconsistent evaluation performance. Cross-corpora has been proposed for better evaluation [124]. Besides, although existing evaluation metric captures grammatical correction and fluency, no one measures the preserving of meaning, which is also necessary to consider when evaluating a GEC system. So, more admirable metrics should explain the grammar, fluency and meaning loyalty of system output.

## VIII. CONCLUSION

We present the first survey in grammar error correction (GEC) for a comprehensive retrospect of existing progress. We first give definition of the task and introduction of public datasets, annotation schema, and two important shared tasks. Then, four dominant basic approaches and their development are explained in detail. After that, we classify the numerous performance boosting techniques into six branches and describe their application and development in GEC, and two data augmentation methods are separately introduced due to their importance. More importantly, in Section 6, after the introduction of the standard evaluation metrics, we give in-depth analysis based on empirical results in aspects of approaches, techniques and integrated GEC systems for a more clear pattern of existing works. Finally, we present five prospective directions based on existing progress in GEC. We hope our effort could provide assistance for future researches in the community.

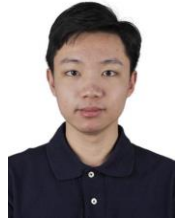

**Yu Wang** is an undergraduate in the College of Artificial Intelligence at Nankai University. His research interests include data-driven approaches to natural language processing and deep reinforcement learning.

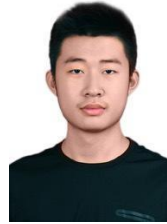

**Yuelin Wang** is an undergraduate in the College of Artificial Intelligence at Nankai University. His research interests in natural language process, deep reinforcement learning and pattern recognition.

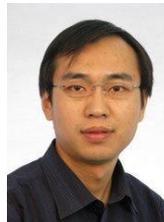

**Jie Liu** is a professor in the College of Artificial Intelligence at Nankai University. His research interests include machine learning, pattern recognition, information retrieval and data mining. He has published several papers on CIKM, ICDM, PAKDD, APWEB, IPM, WWWJ, Soft Computing, etc. He is the coauthor of the best student paper for ICMLC 2013. He has owned the second place in the international ICDAR Book Structure Extraction Competition in 2012 and 2013. He has visited the University of California, Santa Cruz from Sept. 2007 to Sept. 2008. He has visited the Microsoft Research Asia from Aug.2012 to Feb.2013. Prior to joining Nankai University, he obtained his Ph. D. in computer science at Nankai University.

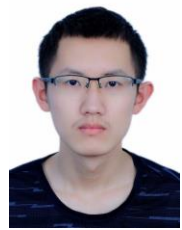

**Zhuo Liu** is an undergraduate in the College of Artificial Intelligence at Nankai University. His research interests include data mining and graph neural network.